\documentclass[10pt,twocolumn,letterpaper]{article}

\usepackage{wacv}
\usepackage{times}
\usepackage{epsfig}
\usepackage{graphicx}
\usepackage{amsmath}
\usepackage{amssymb}
\usepackage{booktabs}       % professional-quality tables
\usepackage{threeparttable}  
\usepackage{paralist}
\usepackage{footnote}
\usepackage{tablefootnote}
\usepackage{verbatim}
\usepackage{comment}
\usepackage{subfig}
\usepackage{caption}
\usepackage{url}     % necessary for citing urls
\usepackage{algorithm}% http://ctan.org/pkg/algorithm
\usepackage{algpseudocode}% http://ctan.org/pkg/algorithmicx
\wacvfinalcopy % *** Uncomment this line for the final submission

\def\wacvPaperID{770} % *** Enter the wacv Paper ID here

\ifwacvfinal\fi
\begin{document}

%%%%%%%%% TITLE
\title{City-Scale Road Extraction from Satellite Imagery v2: \\
Road Speeds and Travel Times}
%\title{Extraction of Road Topology and Travel Times with Satellite Imagery}
\vspace{-5pt}

%RSANE
%RTASE
%CR NASE 
%SIR FATE
%RSATE SI
%\title{CRESSI: City-scale Road Extraction with Speed from Satellite Imagery} % (CRESSI)}
%\title{CERNAS  City-scale Extraction of Road S
% 
%\title{CRESI: City-scale Road Extraction from Satellite Imagery} % (CRESI)}
%Broad Area Road Speed And  BARSA  NE  LARVANE
%broad area road graph ext
%barges
%Larges
%agave
%large area road gragh extraction with Speed
%agave automated graph and velocity extraction

%Optimal routing from satellite imagery (ORSI)
%\title{Automated Road Network from Overhead Imagery (ARNOI)}
%\title{Automated Road Network from Satellite Imagery (ARNSI)}
%\title{Network Extraction from Remote Sensing Imagery (NERSI)}
%\title{Automated Road Network Extraction (ARNE)}
%\title{Overhead Imagery Network Creation {OINC}
%\title{Large Area Satellite Imagery Road Detection (LASIRD)}
%\title{Broad Area Road Network Extraction (BARNE)}
%\title{Broad Area Road Network Detection (BARND)}
%\title{Broad Area Satellite Imagery Road Network Extraction (BASIRNE)}
%\title{Satellite Imagery Road Extraction Network (SIREN)}
%\title{Network Extraction from Satellite Imagery (NESI)}

% Authors at the same institution
%\author{First Author \hspace{2cm} Second Author \\
% Authors at different institutions
\author{Adam Van Etten \\
In-Q-Tel CosmiQ Works\\
{\tt\small avanetten@iqt.org}
}
\vspace{-5pt}

\maketitle
\ifwacvfinal\thispagestyle{empty}\fi

%%%%%%%%% ABSTRACT
\begin{abstract}
Automated road network extraction from remote sensing imagery remains a significant challenge despite its importance in a broad array of applications.  
%To this end, we leverage recent open source advances and the high quality SpaceNet dataset to infer semantic features of the graph, extracting speed limits and route travel times for each roadway. 
To this end, we explore road network extraction at scale with inference of semantic features of the graph, identifying speed limits and route travel times for each roadway.
We call this approach City-Scale Road Extraction from Satellite Imagery v2 (CRESIv2),
%We call this approach City-Scale Road Network Extraction (CRNE).  
% an approach we call City-Scale Road Network Extraction (CRNE).
% City-scale Road Extraction with Speed from Satellite Imagery (CRESSI). 
%To this end, we leverage recent the open source algorithmic advances brought about by the recent SpaceNet road network detection competition to further explore road extraction at scale, 
%Specifically, we scale up SpaceNet algorithms to extract road networks from city-scale regions, an approach we call City-scale Road Extraction from Satellite Imagery (CRESI).  
Including estimates for travel time permits true optimal routing (rather than just the shortest geographic distance), which is not possible with existing remote sensing imagery based methods. %We explore the relative importance of each phase of the our multi-step algorithm with ablation studies, and show that post-processing methods to close road gaps is crucial to performance.  
We evaluate our method using two sources of labels (OpenStreetMap, and those from the SpaceNet dataset), and find that models both trained and tested on SpaceNet labels outperform OpenStreetMap labels by $\geq60\%$.  
%We compare SpaceNet labels to OpenStreetMap (OSM) labels, and find that models both trained and tested on SpaceNet labels outperform OSM labels by $\geq60\%$.  
We quantify the performance of our algorithm with the Average Path Length Similarity (APLS) and map topology (TOPO) graph-theoretic metrics over a diverse  test area covering four cities in the SpaceNet dataset.  For a traditional edge weight of geometric distance, we find an aggregate of 5\% improvement over existing methods for SpaceNet data.  
We also test our algorithm on Google satellite imagery with OpenStreetMap labels, and find a 23\% improvement over previous work.  
Metric scores decrease by only 4\% on large graphs when using travel time rather than geometric distance for edge weights, 
indicating that optimizing routing for travel time is feasible with this approach.  

\end{abstract}

\section{Introduction}\label{sec_intro}

The automated extraction of roads applies to a multitude of long-term efforts: improving access to health services, urban planning, and improving social and economic welfare. This is particularly true in developing countries that have limited resources for manually intensive labeling and are under-represented in current mapping. Updated maps are also crucial for such time sensitive efforts as determining communities in greatest need of aid, effective positioning of logistics hubs, evacuation planning, and rapid response to acute crises.  
%In dynamic scenarios (such as natural disasters) where timely high quality road network revisions are crucial, the manual and semi-automated techniques discussed above often fail to provide updates on the requisite sufficient timescale %(cite puerto rico osm campaign?).  

%Determining optimal routing paths in near real-time is at the heart of many humanitarian, civil, military, and commercial challenges. 
%In the humanitarian realm, for example, traveling to disaster stricken areas can be problematic for relief organizations, particularly if flooding has destroyed bridges or submerged thoroughfares. 
%Autonomous vehicle navigation is one of many examples on the commercial front, as self-driving cars rely heavily upon highly accurate road maps. 

\begin{figure}
\vspace{-5pt}
  \centering
     \includegraphics[width=0.95\linewidth]{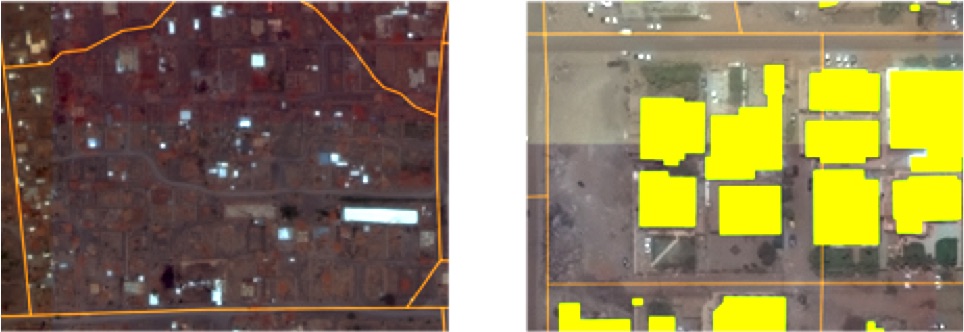}
  \caption{\textbf{Potential issues with OSM data.}  Left: OSM roads (orange) overlaid on Khartoum imagery; the east-west road in the center is erroneously unlabeled.  Right: OSM roads (orange) and SpaceNet buildings (yellow); in some cases road labels are misaligned and intersect buildings.}
   \label{fig:osm_goof}
 \vspace{-15pt}
\end{figure}

Existing data collection methods such as manual road labeling or aggregation of mobile GPS tracks are currently insufficient to properly capture either underserved regions (due to infrequent data collection), or the dynamic changes inherent to road networks in rapidly changing environments.  For example, in many regions of the world OpenStreetMap (OSM) \cite{OpenStreetMap} road networks are remarkably complete.  Yet, in developing nations OSM labels are often missing metadata tags (such as speed limit or number of lanes), or are poorly registered with overhead imagery (i.e., labels are offset from the coordinate system of the imagery), see Figure \ref{fig:osm_goof}. An active community works hard to keep the road network up to date, but such tasks can be challenging and time consuming in the face of large scale disasters. 
For example, following Hurricane Maria, it took the Humanitarian OpenStreetMap Team (HOT) over two months to fully map Puerto Rico \cite{osm_maria}.  
Furthermore, in large-scale disaster response scenarios, pre-existing datasets such as population density and even geographic topology may no longer be accurate, preventing responders from leveraging this data to jump start mapping efforts.

The frequent revisits of satellite imaging constellations may accelerate existing efforts to quickly update road network and routing information.  
Of particular utility is estimating the time it takes to travel various routes in order to minimize response times in various scenarios;
unfortunately existing algorithms based upon remote sensing imagery cannot provide such estimates.
A fully automated approach to road network extraction and travel time estimation from satellite imagery therefore warrants investigation, and is explored in the following sections. 
%Our contributions are as follows.
In Section \ref{sec:existing} we discuss related work, while
Section \ref{sec:algo} details our graph extraction algorithm that infers a road network with semantic features directly from imagery.
In Section \ref{sec:data} we discuss the datasets used and our method for assigning road speed estimates based on road geometry and metadata tags.  
Section \ref{sec:metrics} discusses the need for modified metrics to measure our semantic graph, and 
Section \ref{sec:experiments} covers our experiments to extract road networks from multiple datasets.  
Finally in Sections \ref{sec:discussion} and \ref{sec:conclusion} we discuss our findings and conclusions.
%Section \ref{sec:performance} covers inference results on SpaceNet test chips, as well as comparison to OSM and ablation studies.
%Finally, in Sections \ref{sec:cresi} and \ref{sec:cresi_results}  we describe our methods for scaling to large images, and performance on large test regions.

%The standard structure is: 1. Introduction, 2. Related work, 3. Method, 4. Experiments, 5. Discussion, 6. Conclusion.
%Description of data set and evaluation measures should be moved to a section "Experiments", together with the results. 

\section{Related Work}\label{sec:existing}
%\section{Existing Datasets and Approaches}\label{sec:existing}

Extracting road pixels in small image chips from aerial imagery has a rich history (e.g. 
% {sing18} http://bmvc2018.org/contents/papers/0345.pdf
\cite{zhang2017}, %Zhang et al 2017. %https://arxiv.org/pdf/1711.10684.pdf. 
\cite{mattyus16} % http://www.cs.utoronto.ca/~slwang/hdmap.pdf
\cite{wang2016}, %Wang et al 2016, %http://www.mdpi.com/2220-9964/5/7/114
\cite{zhang2017}, %Zhang et al 2017. %https://arxiv.org/pdf/1711.10684.pdf. 
\cite{sironi14}, %https://icwww.epfl.ch/~lepetit/papers/sironi_cvpr14.pdf
\cite{mnihroads}). %Mnih and Hinton 2010, %http://www.cs.toronto.edu/~fritz/absps/road_detection.pdf
These algorithms typically use a segmentation + post-processing approach combined with lower resolution imagery (resolution $\geq 1$ meter), and OpenStreetMap labels.
Some more recent efforts (e.g. \cite{dlinknet}) have utilized higher resolution imagery (0.5 meter) with pixel-based labels \cite{deepglobe}.  

Extracting road networks directly has also garnered increasing academic interest as of late.  
%been attempted by a number of studies.  
\cite{stoica04} % https://link.springer.com/article/10.1023/B:VISI.0000013086.45688.5d
attempted road extraction via a Gibbs point process, while 
\cite{wegner13} % https://prs.igp.ethz.ch/content/dam/ethz/special-interest/baug/igp/photogrammetry-remote-sensing-dam/documents/pdf/cvpr2013_1227_cr.pdf
% also developed the 5% shortest path metric
showed some success with road network extraction with a conditional random field model. 
\cite{chai13} % https://makerhacker.github.io/paper-mining/cvpr/cvpr2013/cvpr-2013-Recovering_Line-Networks_in_Images_by_Junction-Point_Processes.html
used junction-point processes to recover line networks in both roads and retinal images, while
\cite{turet13}  % https://web.bii.a-star.edu.sg/~zhangxw/files/turetken_tpami_2014.pdf
extracted road networks by representing image data as a graph of potential paths.
\cite{mattyus15} % https://www.cv-foundation.org/openaccess/content_iccv_2015/papers/Mattyus_Enhancing_Road_Maps_ICCV_2015_paper.pdf
extracted road centerlines and widths via OSM and a Markov random field process, and
\cite{mosinska18} %https://arxiv.org/pdf/1712.02190.pdf
 used a topology-aware loss function to extract road networks from aerial features as well as cell membranes in microscopy.  

Of greatest interest for this work are a trio of recent papers that improved upon previous techniques. % to extract road networks from overhead imagery.
%\cite{deeproadmapper}, % http://www.cs.toronto.edu/~wenjie/papers/iccv17/mattyus_etal_iccv17.pdf
%\cite{roadtracer},
DeepRoadMapper \cite{deeproadmapper} used segmentation followed by $A^{*}$ search, applied to the not-yet-released TorontoCity Dataset.  
The RoadTracer paper \cite{roadtracer}
%claimed superior performance to \cite{deeproadmapper}, and 
utilized an interesting approach that used OSM labels to directly extract road networks from imagery without intermediate steps such as segmentation.  While this approach is compelling, according to the authors it ``struggled in areas where roads were close together'' \cite{fbastani_roads} and underperforms other  techniques such as segmentation + post-processing
%(for example, the approach detailed in Section \ref{sec:algo}) 
when applied to higher resolution %SpaceNet 
data with dense labels.   
\cite{Batra_2019_CVPR} used a connectivity task termed Orientation Learning combined with a stacked convolutional module and a SoftIOU loss function to effectively utilize the mutual information between orientation learning and segmentation tasks to extract road networks from satellite imagery, noting improved performance over \cite{roadtracer}.  
Given that \cite{roadtracer} noted superior performance to \cite{deeproadmapper} (as well as previous methods), and \cite{Batra_2019_CVPR} claimed improved performance over both \cite{roadtracer} and \cite{deeproadmapper}, we compare our results to RoadTracer \cite{roadtracer} and Orientation Learning \cite{Batra_2019_CVPR}.

We build upon CRESI v1 \cite{cresi} that scaled up narrow-field road network extraction methods. In this work we focus primarily on developing methodologies to infer road speeds and travel times, but also improve the segmentation, gap mitigation, and graph curation steps of \cite{cresi}, as well as improve inference speed.  
%Given that \cite{roadtracer} noted superior performance to \cite{deeproadmapper} (as well as previous methods), and \cite{Batra_2019_CVPR} claimed improved performance over both \cite{roadtracer} and \cite{deeproadmapper}, in this work we compare our results to \cite{roadtracer} and \cite{Batra_2019_CVPR}.

% In this study we expand upon the work of \cite{cresi} to improve performance and incorporate travel ,,,

% In this study we expand upon the work of \cite{cresi} to improve performance and incorporate travel ,,,

\section{Road Network Extraction Algorithm}\label{sec:algo}

%We initially train and test an algorithm on the SpaceNet $1300 \times 1300$ pixel 
%($\approx160,000 \, \rm{m^2}$) 
%($\approx400 \, \rm{m}$) 
%image chips.
Our approach is to combine novel segmentation approaches, improved post-processing techniques for road vector simplification, and road speed extraction using both vector and raster data.
Our greatest contribution is the inference of road speed and travel time for each road vector, a task that has not been attempted in any of the related works described in Section \ref{sec:existing}.
%Our approach is to combine recent work in satellite imagery semantic segmentation with improved post-processing techniques for road vector simplification.  
%Our approach is to combine at large scale recent work in satellite imagery semantic segmentation and improved post-processing techniques for road vector simplification.  
%We begin with inferring a segmentation mask using convolutional neural networks (CNNs).  
We utilize satellite imagery and geo-coded road centerline labels (see Section \ref{sec:data} for details on datasets) to build training datasets for our models.

We create training masks from road centerline labels assuming a mask halfwidth of 2 meters for each edge.  
While scaling the training mask width with the full width of the road is an option (e.g. a four lane road will have a greater width than a two lane road),
 since the end goal is road centerline vector extraction, we utilize the same training mask width for all roadways. 
 Too wide of a road buffer inhibits the ability to identify the exact centerline of the road, while too thin of a buffer reduces the robustness of the model to noise and variance in label quality; we find that a 2 meter buffer provides the best tradeoff between these two extremes.  

We have two goals: extract the road network over large areas, and assess travel time along each roadway.  
In order to assess travel time we assign a speed limit to each roadway 
%(unless speed limit is explicitly tagged in the available data)  
based on metadata tags such as road type, number of lanes, and surface construction.  
%We subsequently train models according to the segmentation approaches detailed below.

We assign a maximum safe traversal speed of 10 - 65 mph to each segment based on the road metadata tags.  For example, a paved one-lane residential road has a speed limit of 25 mph, a three-lane paved motorway can be traversed at 65 mph, while a one-lane dirt cart track has a traversal speed of 15 mph.  
See Appendix A for further details.  
This approach is tailored to disaster response scenarios, where safe navigation speeds likely supersede government-defined speed limits.
%We are more interested in disaster response scenarios rather than traffic management, so the exact speed limit of each road segment is less important than an estimate of the maximum speed one could safely traverse the segment. 
We therefore prefer estimates based on road metadata over government-defined speed limits, which may be unavailable or inconsistent in many areas.

\subsection{Multi-Class Segmentation}\label{sec:seg_mc}

\begin{figure}[]
\vspace{-5pt}
\begin{center}
\setlength{\tabcolsep}{0.3em}
\begin{tabular}{cc}
\vspace{-6pt}

\subfloat [\textbf{Input Image}] {\includegraphics[width=0.48\linewidth]{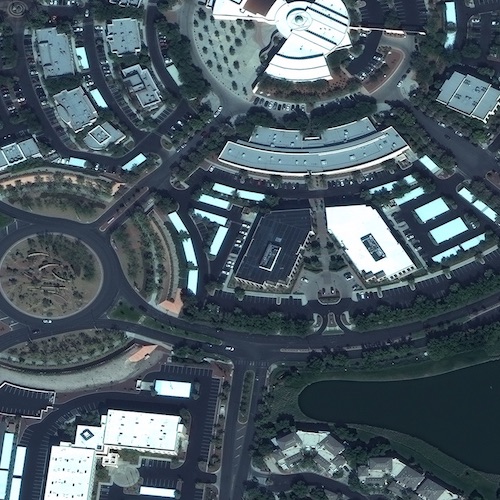}} &
\subfloat [\textbf{Binary Training Mask}] {\includegraphics[width=0.48\linewidth]{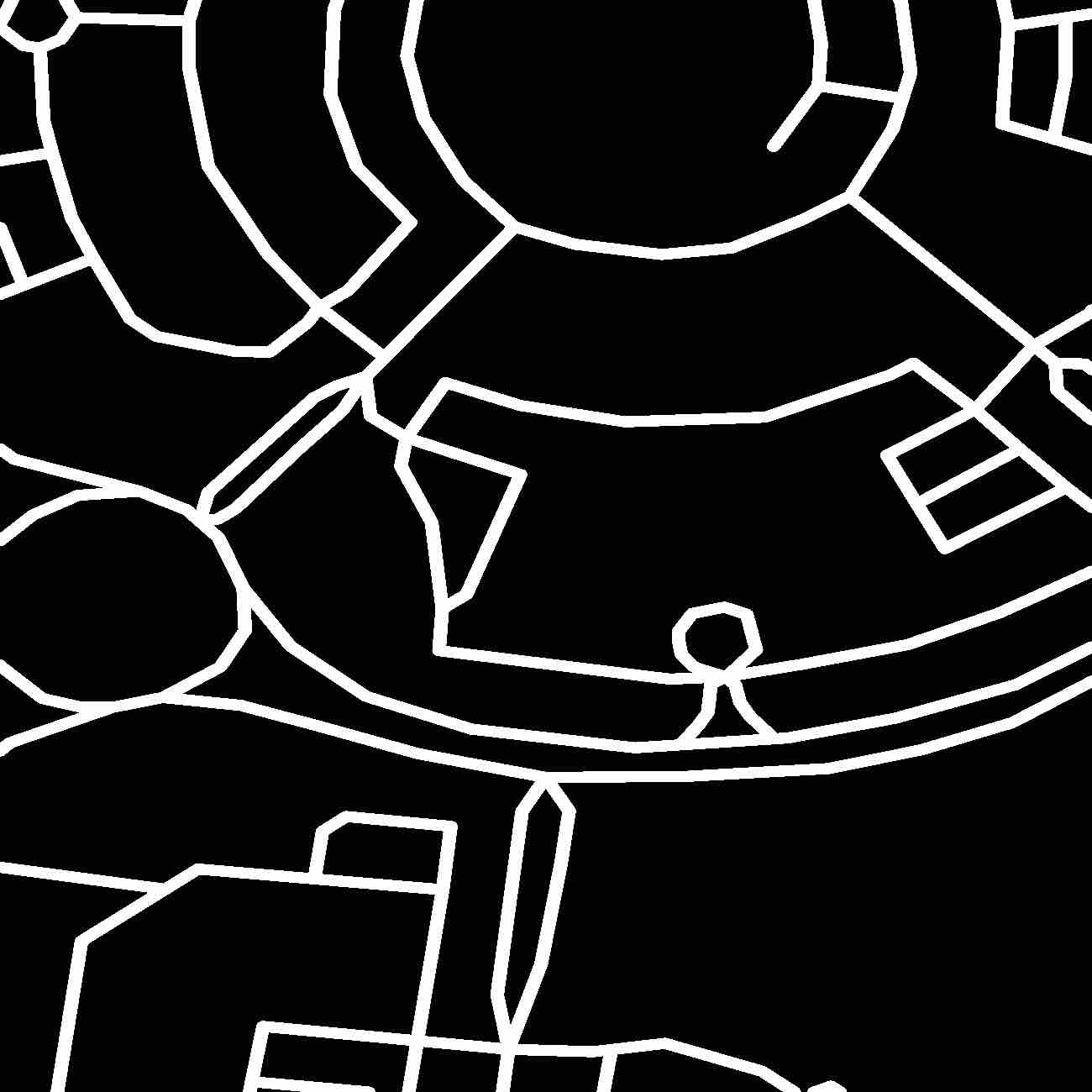}}  \\

\subfloat [\textbf{Continuous Training Mask}]  {\includegraphics[width=0.48\linewidth]{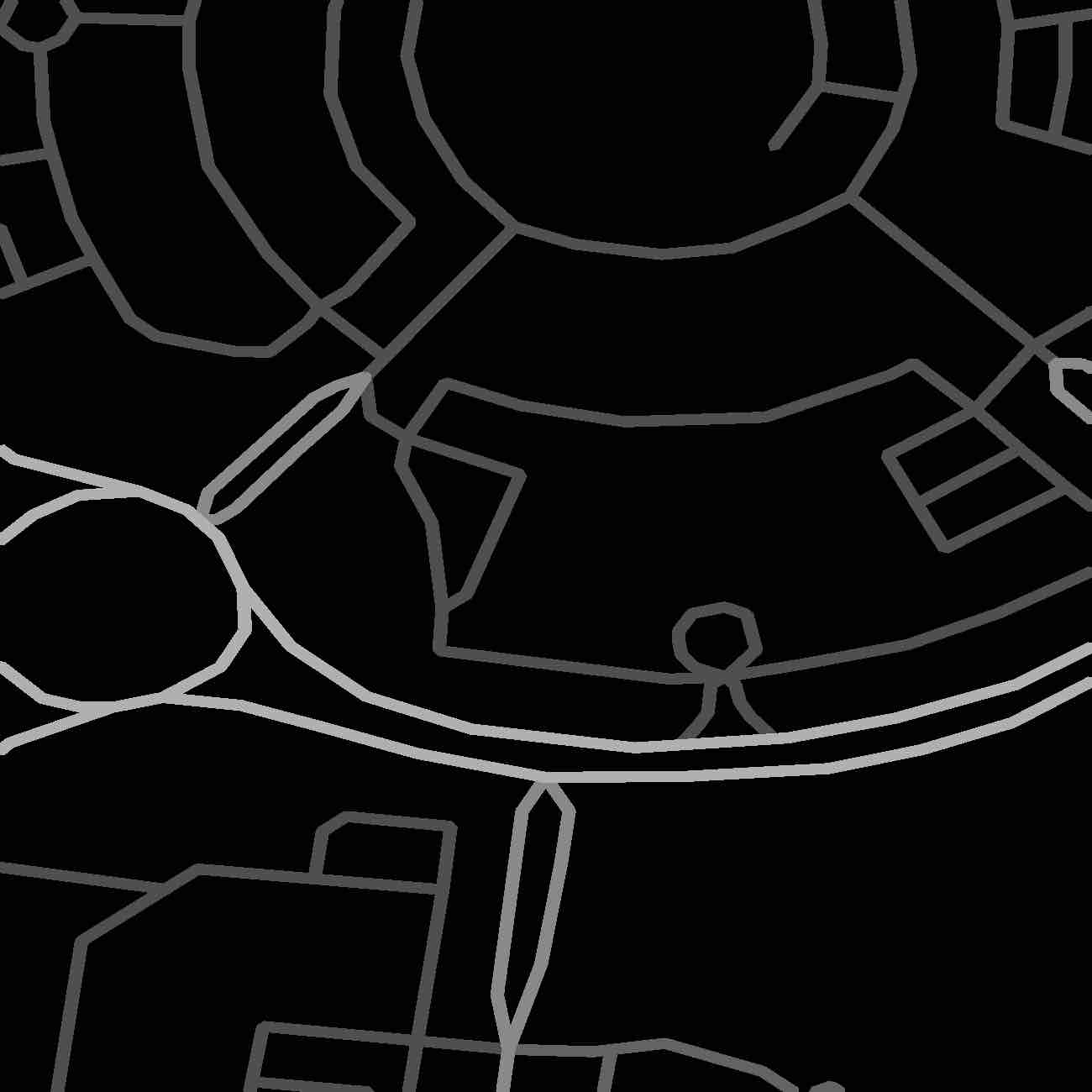}} &
\subfloat [\textbf{Multi-class Training Mask}]  {\includegraphics[width=0.48\linewidth]{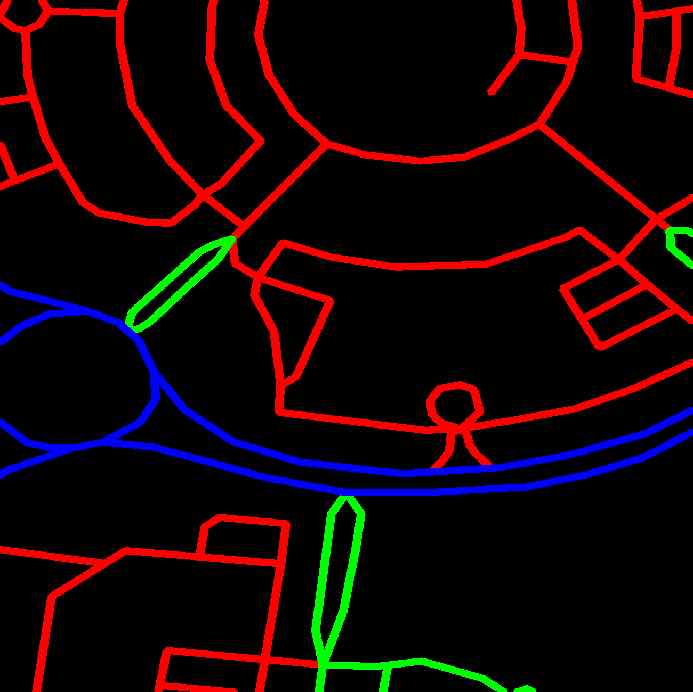}}  \\
\end{tabular}

 \caption{\textbf{Training data.} (a) Input image.  (b) Typical binary road training mask (not used in this study). 
 	(c) Continuous training mask, whiter denotes higher speeds.
	(d) Multi-class mask showing individual speed channels: red = 21-30 mph, green = 31-40 mph, blue = 41-50 mph.}
 \label{fig:train_masks}
\end{center}
%\vspace{-6pt}
\vspace{-12pt}
\end{figure}

We create multi-channel training masks by binning the road labels into a 7-layer stack, with channel 0 detailing speeds between 1-10 mph, channel 1 between 11-20 mph, etc. (see Figure 
\ref{fig:train_masks}).   
%\ref{fig:mc_mask}).   
%We use a similar network architecture to the previous section (ResNet34 encoder with a U-Net inspired decoder), though we use a loss function of: 
We train a segmentation model inspired by the winning SpaceNet 3 algorithm \cite{albu}, and use
a ResNet34 \cite{resnet} encoder with a U-Net \cite{unet} inspired decoder.  We include skip connections every layer of the network, and use an Adam optimizer.
We explore various loss functions, including binary cross entropy, Dice, and focal loss \cite{focal_loss}, and find the best performance with  $\alpha_{mc} = 0.75$ and 
a custom loss function of:
\begin{equation}\label{eqn:c}
%C= 0.75 \times {\rm focal} \, + \,0.25 \times (1 - {\rm dice})
%%C= 0.8 \times {\rm binary\_cross\_entropy} \, + \,0.2 \times (1 - {\rm dice\_coeff}).
    \mathcal{L} = \alpha_{mc}\mathcal{L}_{\text{focal}} + (1-\alpha_{mc})\mathcal{L}_{\text{dice}} 
\end{equation}
%where `focal' is focal loss \cite{focal_loss}, `dice' is the Dice coefficient, and $\alpha_{mc} = 0.75.$ 
%is a constant.% = 0.75$.
%As before, training data is split into four folds and training occurs for 30 epochs. 
% At inference time the folds are merged by mean to give the final road mask prediction.

% combined figure with mult-class and continuous masks

%%%%%%%%%%%
% if we don't want to include continuous masks
\begin{comment}
\begin{figure}[]
\vspace{-5pt}
\begin{center}
\setlength{\tabcolsep}{0.3em}
\begin{tabular}{cc}
\vspace{-5pt}

\subfloat{\includegraphics[width=0.48\linewidth]{RGB-PanSharpen_AOI_2_Vegas_img47_mask_black.jpg}}  &
\subfloat{\includegraphics[width=0.48\linewidth]{mc_train_mask1p5.jpg}} 
\end{tabular}

 \caption{\textbf{Multi-class training mask.} Left: Typical binary road training mask.  Right: Multi-class training mask showing individual speed channels: red = 21-30 mph, green = 31-40 mph, blue = 41-50 mph.}
 \label{fig:mc_mask}
\end{center}
\vspace{-6pt}
\end{figure}
\end{comment}
%%%%%%%%%%

%\begin{comment}
\subsection{Continuous Mask Segmentation}

A second segmentation method renders continuous training masks from the road speed labels.  Rather than the typical binary mask, we linearly scale the mask pixel value
% [0 .. 255] 
with speed limit, 
assuming a maximum speed of 65 mph 
% with a maximum pixel value of 255 (for 65 mph) %, and a minimum burn value of 40 for the rare 15 mph roads 
(see Figure \ref{fig:train_masks}).
%(see Figure \ref{fig:mask_contin}).

%%%%%%%%
\begin{comment}
\begin{figure}[]
\vspace{-5pt}
\begin{center}
\setlength{\tabcolsep}{0.3em}
\begin{tabular}{cc}
\vspace{-5pt}x

\subfloat{\includegraphics[width=0.48\linewidth]{RGB-PanSharpen_AOI_2_Vegas_img47.jpg}} &
\subfloat{\includegraphics[width=0.48\linewidth]{RGB-PanSharpen_AOI_2_Vegas_img47_mask.jpg}} \\
\end{tabular}

 \caption{\textbf{Continuous training mask.} Left: Sample SpaceNet image chip.  Right: Continuous training mask, whiter denotes higher speeds.}
 \label{fig:mask_contin}
\end{center}
\vspace{-6pt}
\end{figure}
\end{comment}
%%%%%%%%%

We use a similar network architecture to the previous section (ResNet34 encoder with a U-Net inspired decoder), though we use a loss function that utilizes cross entropy (CE) rather than focal loss ($\alpha_c = 0.75$): 
%We train a segmentation model inspired by the winning SpaceNet 3 algorithm \cite{albu}, and use
%a ResNet34 \cite{resnet} encoder with a U-Net \cite{unet} inspired decoder.  We include skip connections every layer of the network, with an Adam optimizer and a custom loss function of:
\begin{equation}
\label{eqn:c2}
    \mathcal{L} = \alpha_c\mathcal{L}_{\text{CE}} + (1-\alpha_c)\mathcal{L}_{\text{dice}}
%C= 0.8 \times {\rm CE} \, + \,0.2 \times (1 - {\rm dice}).
\end{equation}
%where `CE' is cross entropy, `dice' is the Dice coefficient, and $\alpha_c = 0.75.$ 
% is a constant. %=0.8$.
%Training data is split into four folds and training occurs for 30 epochs.  At inference time the folds are merged by mean to give the final road mask prediction.
%\end{comment}

\subsection{Graph Extraction Procedure}

\begin{figure}[]
\vspace{-5pt}
  \centering
     \includegraphics[width=0.95\linewidth]{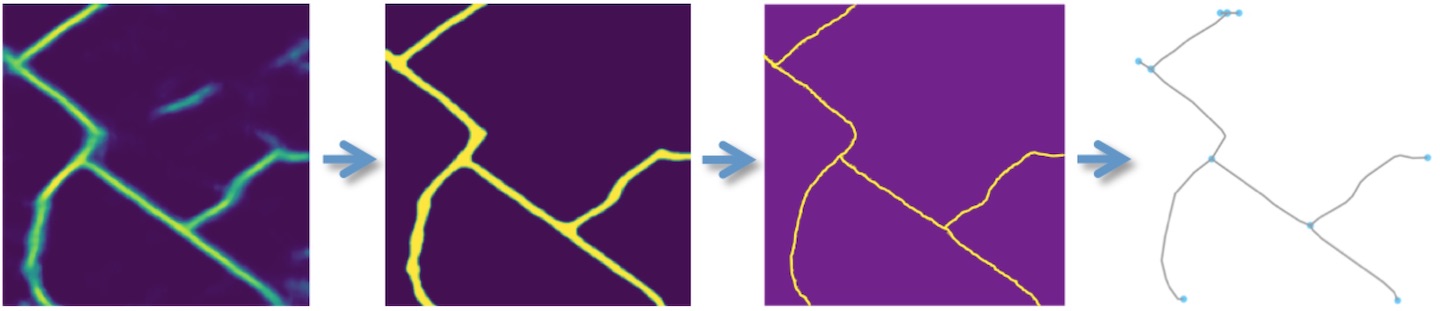}
  \caption{\textbf{Graph extraction procedure.}  Left: raw mask output.
  %Using road masks, we train CNN segmentation algorithms (such as PSPNet \cite{pspnet} or U-Net \cite{unet}) to infer road masks from SpaceNet imagery.  
  Left center: refined mask.  Right center: mask skeleton.
  %  \footnote{https://scikit-image.org/docs/dev/auto_examples/edges/plot_skeleton.html})
  Right: graph structure.  
  %Competitors to the roads challenge generally followed the same approach as this baseline algorithm, albeit with different network architectures and more refined post-processing steps.
}
    \label{fig:baseline}
\vspace{-8pt}
\end{figure}

The output of the segmentation mask step detailed above is subsequently refined into road vectors.  
We begin by smoothing the output mask with a Gaussian kernel of 2 meters. %
% and flattening the final output mask to create a binary prediction mask.
This mask is then refined using opening and closing techniques with a similar kernel size of 2 meters, 
as well as removing small object artifacts or holes with an area less than 30 square meters.  
From this refined mask
we create a skeleton (e.g.~ sckit-image skeletonize \cite{skimage}).
This skeleton is rendered into a graph structure with a version of 
the {\it sknw} package \cite{sknw} package modified to work on very large images.
This process is detailed in Figure \ref{fig:baseline}. 
The graph created by this process contains length information for each edge, but no other metadata.  
To close small gaps and remove spurious connections not already corrected by the mask refinement procedures, 
we remove disconnected subgraphs with an integrated path length of less than
% 80 meters.  
 a certain length (6 meters for small image chips, and 80 meters for city-scale images).
 We also follow \cite{albu} and remove terminal vertices that lie on an edge less than 3 meters in length, 
 and connect terminal vertices if the distance to the nearest non-connected node is less than 6 meters.  

%\begin{figure}[]
%  \centering
%     \includegraphics[width=0.95\linewidth]{baseline_ex.jpg}
%  \caption{Sample output of baseline algorithm applied to SpaceNet test data.}
%    \label{fig:baseline_ex}
%\end{figure}

%%%%%%%%%%%%%%%%%%%%
% mask + network in shanghai, not great predictions
\begin{comment}
\begin{figure}[]
  \begin{subfigure}[b]{0.48\columnwidth}
    \includegraphics[width=\linewidth]{speed0.jpg}
 \caption{}
 %   \label{fig:1}
  \end{subfigure}
	\hfill %
  \begin{subfigure}[b]{0.48\columnwidth}
    \includegraphics[width=\linewidth]{speed1.jpg}
  \caption{}
  %  \label{fig:2}
  \end{subfigure}
 	 \hfill %
 \vspace{-2mm}%Put here to reduce too much white space after your table 
 \caption{(a) Flattened prediction mask for SpaceNet test chip.  (b) Imagery overlaid with inferred network (yellow).}
 \label{fig:mc_mask2}
 %\vspace{-4mm}%Put here to reduce too much white space after your table 
\end{figure}
\end{comment}
%%%%%%%%%%%%%%%%%%%%%

\subsection{Speed Estimation Procedure}\label{sec:speed_ex}

\begin{figure}[t]
\vspace{-1pt}
  \centering
     \includegraphics[width=0.98\linewidth]{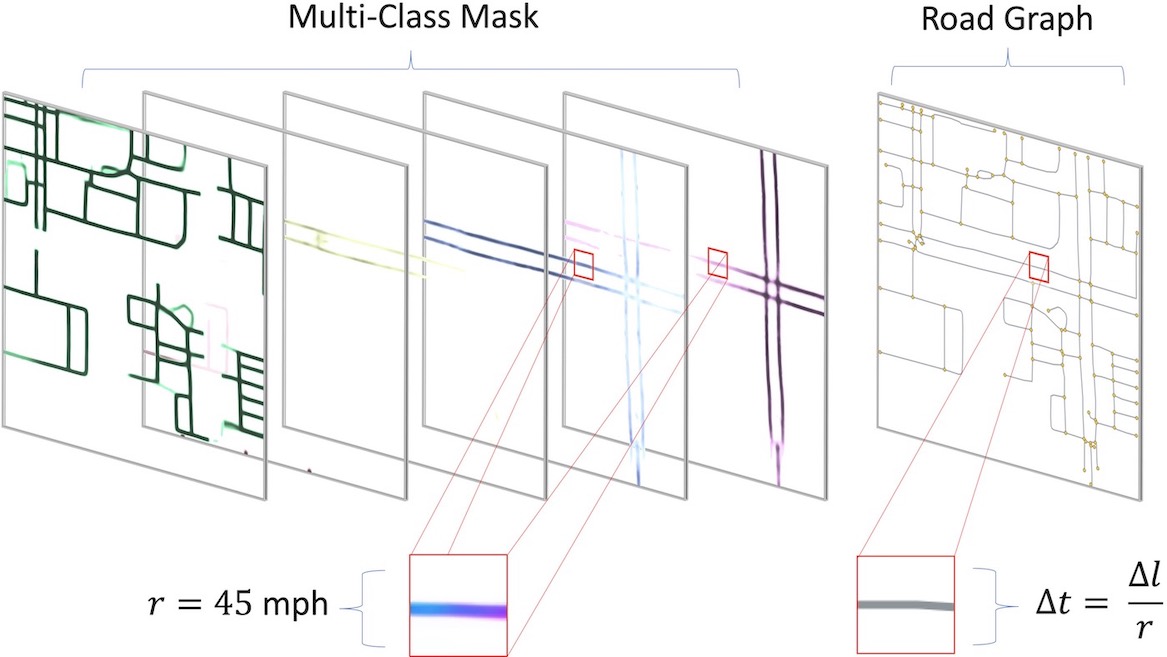}
  \caption{\textbf{Speed estimation procedure.} 
  Left: Sample multi-class prediction mask; the speed ($r$) of an individual patch (red square) can be inferred by measuring the signal from each channel.  
  Right: Computed road graph; travel time ($\Delta t$) is given by speed ($r$) and segment length ($\Delta l$).}
    \label{fig:speed_comp}
\vspace{-8pt}
\end{figure}

We estimate travel time for a given road edge by leveraging the speed information encapsulated in the prediction mask.  
The majority of edges in the graph are composed of multiple segments; accordingly, we attempt to estimate the speed of each segment in order to determine the mean speed of the edge.  
%The majority of edges in the graph are composed of multiple segments (e.g. the edge in Figure \ref{fig:speed_comp} has 6 segments).  
%Accordingly, we attempt to estimate the speed of each segment in order to determine the mean speed of the edge.  
This is accomplished by analyzing the prediction mask at the location of the segment midpoints.  For each segment in the edge, at the location of the midpoint of the segment we extract a small $8\times8$ pixel patch from the prediction mask.   The speed of the patch is estimated by filtering out low probability values (likely background), and averaging the remaining pixels (see Figure \ref{fig:speed_comp}).  
In the multi-class case, if the majority of the the high confidence pixels in the prediction mask patch belong to channel 3 (corresponding to 31-40 mph), we would assign the speed at that patch to be 35 mph.  
For the continuous case the inferred speed is simply directly proportional to the mean pixel value.
%We estimate speed for each edge by extracting patches from the training mask at the location of nodes of interest.  Edges in the graph contain geometric information, and the points of this geometry can be queried for the local speed estimate.  At each segment in the graph edge we insert a midpoint to estimate the speed of that segment.  We  then extract the mask patch ($4\times4$ pixels be default) at the location of the midpoint (see Figure \ref{fig:speed_comp}).  To estimate the actual speed at the patch location we take the mean of the 90th percentile and above for the flattened patch. This ensures that background or low confidence predictions do not skew the speed estimate.  

%%%%%%%%%%%%%%%%%%%%%
\begin{comment}
\begin{figure}[t]
\vspace{-5pt}
  \centering
     \includegraphics[width=0.95\linewidth]{speed_comp2.jpg}
  \caption{\textbf{Segment speed estimation.} Top: Example road edge (yellow), with midpoints of each segment shown in blue. Bottom left: Zoomed in region around the green midpoint.  Bottom right: Corresponding portion of the prediction mask ($8\times8$ pixel patch size in red); speed is estimated with this mask patch. }
    \label{fig:speed_comp}
\vspace{-6pt}
\end{figure}
\end{comment}
%%%%%%%%%%%%%%%%%%%%%

The travel time for each edge is in theory the path integral of the speed estimates at each location along the edge.  But given that each roadway edge is presumed to have a constant speed limit, we refrain from computing the path integral along the graph edge.  Instead, we estimate the speed limit of the entire edge by taking the mean of the speeds at each segment midpoint.  Travel time is then calculated as edge length divided by mean speed.  
%For the continuous mask case this is achieved by taking the mean of the speeds at each segment midpoint.  
%For the multi-channel mask algorithm we assign the speed limit as the maximum channel among all the edge segments

%%%%%%%%%%%%%%%%%%%%%
\begin{comment}

The inference algorithm is summarized in Table \ref{tab:algo}.

\begin{table}[h]
  \caption{Graph inference algorithm}
  \vspace{-3pt}
  \label{tab:algo}
  \small
  \centering
   \begin{tabular}{ll}
    \toprule
     Step & Description \\
    \toprule
%	1 & Apply the 4 trained segmentation models to test data \\
%	2 & Merge these 4 predictions into a total road mask \\
	1 & Compute road mask from segmentation model \\
	2 & Clean road mask with opening, closing, smoothing \\ %procedures 
	3 & Skeletonize road mask \\ % with {\it scikit-image.skelotonize} \\
	4 & Extract graph from skeleton \\ % with {\it sknw} \\
	5 & Remove spurious edges and close small gaps in graph \\
	6 & Estimate local speed limit at midpoint of each segment \\
	7 & Assign travel time to each edge from aggregate speed \\
    \bottomrule
  \end{tabular}
\vspace{-8pt}
\end{table}

%\begin{enumerate}
%\item Apply the 4 trained segmentation models to test data
%\item Merge these 4 predictions into a total road mask
%\item Clean road mask with opening, closing, smoothing %procedures
%\item Skeletonize road mask with {\it scikit-image.skelotonize}
%\item Extract graph from skeleton with {\it sknw}
%\item Remove spurious edges and close small gaps in graph
%\end{enumerate}

\end{comment}
%%%%%%%%%%%%%%%%%%%%%

\subsection{Scaling to Large Images}\label{sec:CRESIv2}

\begin{figure}[]
%\vspace{-5pt}
\centering
\includegraphics[width=0.98\linewidth]{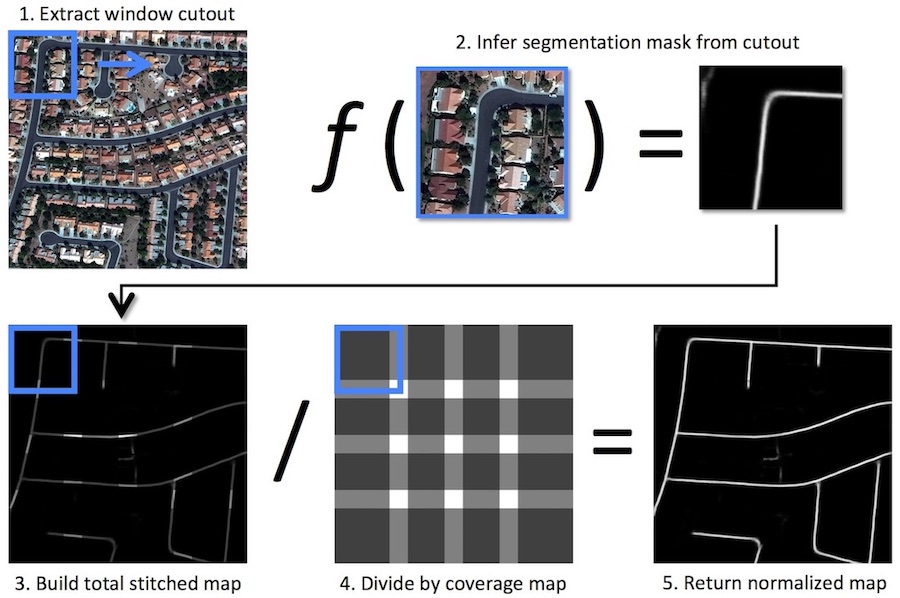}
\caption{\textbf{Large image segmentation.} 
BASISS process of segmenting an arbitarily large satellite image \cite{basiss}.}
%BASISS process of slicing a large satellite image (top) and ground truth road mask (bottom) into smaller cutouts for segmentation training or inference \cite{basiss}.}
\label{fig:SIMRDWN_training}
\vspace{-6pt}
\end{figure}

The process detailed above
%in Section \ref{sec:algo} 
works well for small input images,
%below $\sim2000 \times 2000$ pixels in extent, 
yet fails for 
%images larger than this 
large images due to a saturation of GPU memory.  For example, even for a relatively simple architecture such as U-Net \cite{unet}, typical GPU hardware (NVIDIA Titan X GPU with 12 GB memory) will saturate for images greater than $\sim2000 \times 2000$ pixels in extent and reasonable batch sizes $> 4$. 
In this section we describe a straightforward methodology for scaling up the algorithm to larger images.  
We call this approach City-Scale Road Extraction from Satellite Imagery v2 (CRESIv2).
%We call this approach City-Scale Road Network Extraction (CRNE).
%City-scale Road Extraction with Speed from Satellite Imagery (CRESSI).
%Network Extraction from Satellite Imagery (NESI).  
The essence of this approach is to combine the approach of Sections \ref{sec:seg_mc} - \ref{sec:speed_ex} with the 
Broad Area Satellite Imagery Semantic Segmentation (BASISS) \cite{basiss} methodology.  BASISS returns a road pixel mask for an arbitrarily large test image (see Figure \ref{fig:SIMRDWN_training}), which we then leverage into an arbitrarily large graph.
%The steps of BASISS are outlined in Figure \ref{fig:SIMRDWN_training}.
%The first step in this methodology is provided  by the Broad Area Satellite Imagery Semantic Segmentation (BASISS) \cite{basiss} methodology; this approach is outlined in Figure \ref{fig:SIMRDWN_training} and returns a road pixel mask for a large test image.

%\begin{figure*}[h]

%We take this scaled road mask and apply Steps 2-7 of Table \ref{tab:algo} to retrieve the final road network prediction.  
The final algorithm is given by Table \ref{tab:algo2}.  
%The CRNE algorithm outputs a {\it networkx} \cite{networkx} graph structure, with full access to the 
The output of the CRESIv2 algorithm is a {NetworkX} \cite{networkx} graph structure, with full access to the many algorithms included in this package.

\begin{table}[]
  \caption{CRESIv2 Inference Algorithm}
  \vspace{-3pt}
  \label{tab:algo2}
  \small
  \centering
   \begin{tabular}{ll}
    \toprule
     Step & Description \\
    \toprule
    	1 & Split large test image into smaller windows \\
	2 & Apply multi-class segmentation model to each window \\
	$\,\,2_{\rm b}$ & *\,\,Apply remaining (3) cross-validation models \\
	$\,\,2_{\rm c}$ & *\,\,For each window, merge the 4 predictions \\
%	2 & Apply the 4 trained models to each window \\
%	3 & For each window, merge the 4 predictions \\
	3 & Stitch together the total normalized road mask \\
	4 & Clean road mask with opening, closing, smoothing \\ %procedures 
	5 & Skeletonize flattened road mask \\ % with {\it scikit-image.skelotonize} \\
	6 & Extract graph from skeleton \\ % with {\it sknw} \\
	7 & Remove spurious edges and close small gaps in graph \\
	8 & Estimate local speed limit at midpoint of each segment \\
	9 & Assign travel time to each edge from aggregate speed \\	
	%5 & Clean road mask with opening, closing, smoothing \\ %procedures 
	%6 & Skeletonize road mask \\ %with {\it scikit-image.skelotonize} \\
	%7 & Extract graph from skeleton \\%with {\it sknw} \\
	%8 & Remove spurious edges and close small gaps in graph \\
    \bottomrule
    & * Optional
  \end{tabular}
   \vspace{-5pt} 
\end{table}

%\section{Experiments}
\section{Datasets}\label{sec:data}

Many existing publicly available labeled overhead or satellite imagery datasets tend to be relatively small, or labeled with lower fidelity than desired for foundational mapping.  
%OSM provides road centerlines for much of the world's road networks, though Metadata tags are often available in OSM data, though tags are frequently missing and inconsistently labeled
For example, the International Society for Photogrammetry and Remote Sensing (ISPRS) semantic labeling benchmark \cite{isprs_sem}
%\footnote{http://www2.isprs.org/commissions/comm3/wg4/semantic-labeling.html} 
dataset contains high quality 2D semantic labels over two cities in Germany; 
% Vaihingen = 9cm resolution, 33 patches  The Vaihingen data is composed by a total of 33 image tiles (average size of 2494 × 2064), 16 of which are fully annotated (https://arxiv.org/pdf/1608.00775.pdf), area = 1.4 km2
% Potsdam: 38 tiles of size 6000×6000 pixels, with a spatial resolution of 5cm, area = 3.4 km2
imagery is obtained via an aerial platform and is 3 or 4 channel and  5-10 cm in resolution, though covers only 4.8 km$^2$.  The TorontoCity Dataset \cite{torontocity} contains high resolution 5-10 cm aerial 4-channel imagery, and $\sim700$ km$^2$ of coverage; building and roads are labeled at high fidelity (among other items), but the data has yet to be publicly released.  The Massachusetts Roads Dataset \cite{MnihThesis} contains 3-channel imagery at 1 meter resolution, and $2600$ km$^2$ of coverage; the imagery and labels are publicly available, though labels are scraped from OpenStreetMap and not independently collected or validated. 
% https://medium.com/the-downlinq/broad-area-satellite-imagery-semantic-segmentation-basiss-4a7ea2c8466f
The large dataset size, higher 0.3 m resolution, and hand-labeled and quality controlled labels of SpaceNet \cite{spacenet} provide 
%a significant enhancement over current datasets and provide 
an opportunity for algorithm improvement.  In addition to road centerlines, the SpaceNet dataset contains metadata tags for each roadway including: number of lanes, road type (e.g.~ motorway, residential, etc), road surface type (paved, unpaved), and bridgeway (true/false).  

\subsection{SpaceNet Data}
\begin{figure}[]
\vspace{-5pt}
  \centering
     \includegraphics[width=0.95\linewidth]{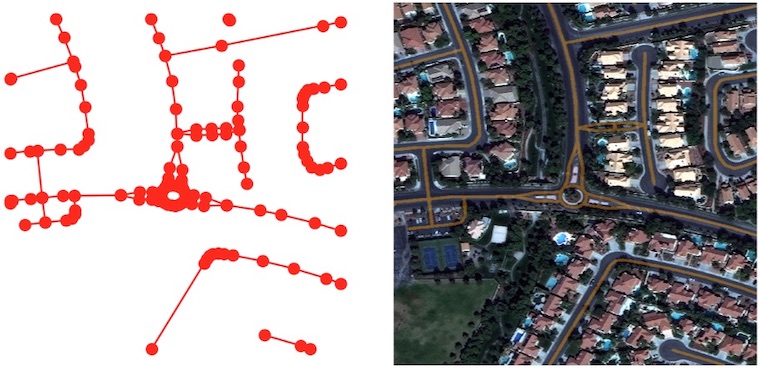}
  \caption{\textbf{SpaceNet training chip.}  Left: SpaceNet GeoJSON road label. Right: $400 \times 400$ meter image overlaid with road centerline labels (orange).}
  \label{fig:sn_data1}
\vspace{-10pt}
\end{figure}

Our primary dataset accordingly consists of the SpaceNet 3  WorldView-3 DigitalGlobe satellite imagery (30 cm/pixel) and attendant road centerline labels.  Imagery covers 3000 square kilometers, and over 8000 km of roads are labeled \cite{spacenet}.  
%Semantic features are labeled for each road segment: number of lanes, road type (e.g. residential, primary, motorway), and road surface type (e.g. dirt, paved). 
Training images and labels are tiled into $1300 \times 1300$ pixel 
%($\approx400 \, \rm{m}$) 
($\approx160,000 \, \rm{m^2}$) 
chips (see Figure \ref{fig:sn_data1}).
%This imagery was used to run a public competition hosted on TopCoder: The SpaceNet Road Detection and Routing Challenge.

\begin{comment}
We assign a maximum safe traversal speed of 10 - 65 mph to each segment based on the road metadata tags.  For example, a paved one-lane residential road has a speed limit of 25 mph, a three-lane paved motorway can be traversed at 65 mph, while a one-lane dirt cart track has a traversal speed of 15 mph.  
See Appendix A for further details.  
This approach is tailored to disaster response scenarios, where safe navigation speeds likely supersede government-defined speed limits.
%We are more interested in disaster response scenarios rather than traffic management, so the exact speed limit of each road segment is less important than an estimate of the maximum speed one could safely traverse the segment. 
We therefore prefer estimates based on road metadata over government-defined speed limits, which may be unavailable or inconsistent in many areas.  
\end{comment}

To test the city-scale nature of our algorithm, we extract large test images from all four of the SpaceNet cities with road labels: Las Vegas, Khartoum, Paris, and Shanghai.  
As the labeled SpaceNet test regions are non-contiguous and irregularly shaped,
%(see Figure \ref{fig:shanghai_reg0}), 
we define rectangular subregions of the images where labels do exist within the entirety of the region.  
These test  regions total 608 km$^2$, with a total road length of 9065 km. See Appendix B for further details.

\subsection{Google / OSM Dataset}

We also evaluate performance with the satellite imagery corpus used by \cite{roadtracer}.  This dataset consists of 
Google satellite imagery at 60 cm/pixel over 40 cities, 25 for training and 15 for testing.  Vector labels are scraped from 
OSM, and we use these labels to build training masks according the procedures described above.  Due to the high variability in OSM road
metadata density and quality, we refrain from inferring road speed from this dataset, and instead leave this for future work.  

\section{Evaluation Metrics}\label{sec:metrics}

%The third SpaceNet competition aimed to extract road networks from satellite imagery.  
Historically, pixel-based metrics (such as IOU or F1 score) have been used to assess the quality of road proposals, though such metrics are suboptimal for a number of reasons (see \cite{spacenet} for further discussion).  Accordingly, we use the graph-theoretic Average Path Length Similarity (APLS) and map topology (TOPO) \cite{topo_metric} metrics designed to measure the similarity between ground truth and proposal graphs. 
%To this end, the graph-theoretic Average Path Length Similarity (APLS) \cite{apls} metric was created to measure similarity between ground truth and proposal road graphs. 

\subsection{APLS Metric}

%Our focus on road networks therefore naturally leads us to consider graph theory as a means to evaluate proposals. While many proposals exist for graph similarity matching, these tend to focus on the logical topology (connections between nodes) of the graph, whereas we care about both logical topology as well as the physical topology of the roads. Since we are primarily interested in routing, we propose a graph theoretic metric based upon Dijkstra's shortest path algorithm \cite{Dijkstra1959}. In essence, we sum the differences in optimal paths between ground truth and proposal graphs.  
%This is illustrated in Figure \ref{fig:apls_fig0}.  

%%%%%%%%
\begin{comment}
\begin{figure}
%\vspace{-5pt}
  \centering
     \includegraphics[width=0.95\linewidth]{apls_fig0.jpg}
  \caption{\textbf{APLS metric.} APLS compares path length difference between sample ground truth and proposal graphs. Left: Shortest path between source (green) and target (red) node in the ground truth graph is shown in yellow, with a path length of $\approx948$ meters. Right: Shortest path between source and target node in the proposal graph with 30 edges removed, with a path length of $\approx1027$ meters; this difference in length forms the basis for the metric.
%  graph similarity metric. 
  %Plotting is accomplished via the {\it osmnx} python package \cite{osmnx}.% \footnote{https://github.com/gboeing/osmnx}. 
  }
  \label{fig:apls_fig0}
\vspace{-6pt}
\end{figure}
\end{comment}
%%%%%%%%

To measure the difference between ground truth and proposal graphs,  the 
 APLS \cite{spacenet}
 %Average Path Length Similarity (APLS) 
 metric sums the differences in optimal path lengths between nodes in the ground truth graph G and the proposal graph G',
 %In effect, this metric repeats the path difference calculation shown in Figure \ref{fig:apls_fig0} for all paths in the graph
 %(plotting is accomplished via the {\it osmnx} python package \cite{osmnx}).% \footnote{https://github.com/gboeing/osmnx}
 with missing paths in the graph assigned a score of 0.
 %Missing paths in the graph are assigned the maximum proportional difference of 1.0. 
 The APLS metric scales from 0 (poor) to 1 (perfect).
%Inherent to this metric is the notion of betweenness centrality, or the number of times a node appears along the shortest paths between other nodes. 
Missing nodes of high centrality will be penalized much more heavily by the APLS metric than missing nodes of low centrality. 
%This feature is intentional, as heavily trafficked intersections are much more important than cul-de-sacs for routing purposes. 
%Any proposed graph G' with missing edges (e.g.~if an overhanging tree is inferred to sever a road) will be heavily penalized by the APLS metric, so ensuring that roads are properly connected is crucial for a high score. 
%For small graphs the greedy approach of computing all possible paths in the network is entirely feasible (computing all possible paths for the 400 m image chips of SpaceNet Challenge 3 takes less than 1 millisecond). For larger graphs, one must decide which nodes and paths are of paramount interest, lest the computational load become burdensome.  For graphs of greater than 1000 nodes, we refrain from injecting midpoints along edges, and select 500 random control nodes.
%; this approach yields an APLS score $\approx7\%$ lower than the greedy approach since midpoints along the graph edges tend to increase the total score.  
%See \cite{spacenet} for further details on the metric.
% The use of path lengths as the core concept of APLS increases the flexibility of the metric, as the definition of shortest path can be user defined. 
The definition of shortest path can be user defined; the natural first step is to consider geographic distance as the measure of path length (APLS$_{\rm{length}}$), but any edge weights can be selected.  Therefore, if we assign a travel time estimate to each graph edge we can use the APLS$_{\rm{time}}$ metric to measure differences in travel times between ground truth and proposal graphs.
%The use of path lengths as the core concept of APLS greatly enhances the flexibility of the metric.  The metric can be applied to non-imagery data, and even more importantly, the definition of shortest path can be user defined.  The natural first step is to consider geographic distance as the measure of path length (APLS$_{\rm{length}}$), but any edge weights can be selected.  Therefore, if we assign a travel time estimate to each graph edge we can use the APLS$_{\rm{time}}$ metric to measure differences in travel times between ground truth and proposal graphs.

For large area testing, evaluation takes place with the APLS metric adapted for large images: no midpoints along edges and a maximum of 500 random control nodes.

\subsection{TOPO Metric}

The TOPO metric \cite{topo_metric} is an alternative metric for computing road graph similarity.  TOPO compares the nodes that can be reached within a small local vicinity of a number of seed nodes, categorizing proposal nodes as true positives, false positives, or false negatives depending on whether they fall within a buffer region (referred to as the ``hole size''). By design, this metric evaluates local subgraphs in a small subregion ($\sim 300$ meters in extent), and relies upon physical geometry.  Connections between greatly disparate points ($>300$ meters apart) are not measured, and the reliance upon physical geometry means that travel time estimates cannot be compared.

\section{Experiments}\label{sec:experiments}

We train CRESIv2 models on both the SpaceNet and Google/OSM datasets.  For the SpaceNet models, we use the 2780 images/labels in the SpaceNet 3 training dataset.  The Google/OSM models are trained with the 25 training cities in \cite{roadtracer}.  All segmentation models use a road centerline halfwidth of 2 meters, and withhold 25\% of the training data for validation purposes.  Training occurs for 30 epochs.  Optionally, one can create an ensemble of 4 folds (i.e. the 4 possible unique combinations of 75\% train and 25\% validate) to train 4 different models.  This approach may increase model robustness, at the cost of increased compute time.  
%at the cost of $4\times$ the compute requirements.  
% If 4 folds are used, at inference time the final prediction mask is simply the mean of the 4 prediction masks.
As inference speed is a priority, all results shown below use a single model, rather than the ensemble approach.  
%Training data is split into four folds and training occurs for 30 epochs.  At inference time the folds are merged by mean to give the final road mask prediction.
%For both segmentation methods, training data is split into four folds and training occurs for 30 epochs.  
%At inference time each model is used for prediction and final mask is determined by averaging.
% We remove disconnected subgraphs with an integrated path length of less than 80 meters in length.

For the Google / OSM data, we train a segmentation model as in Section \ref{sec:seg_mc}, though with only a single class since we forego speed estimates with this dataset.
%\section{Algorithm Performance}\label{sec:performance}

\subsection{SpaceNet Test Corpus Results}

\begin{figure}[]
\vspace{-5pt}
\begin{center}
\setlength{\tabcolsep}{0.3em}
\begin{tabular}{cc}
\vspace{-5pt}

\subfloat [\textbf{Ground truth mask}] {\includegraphics[width=0.48\linewidth]{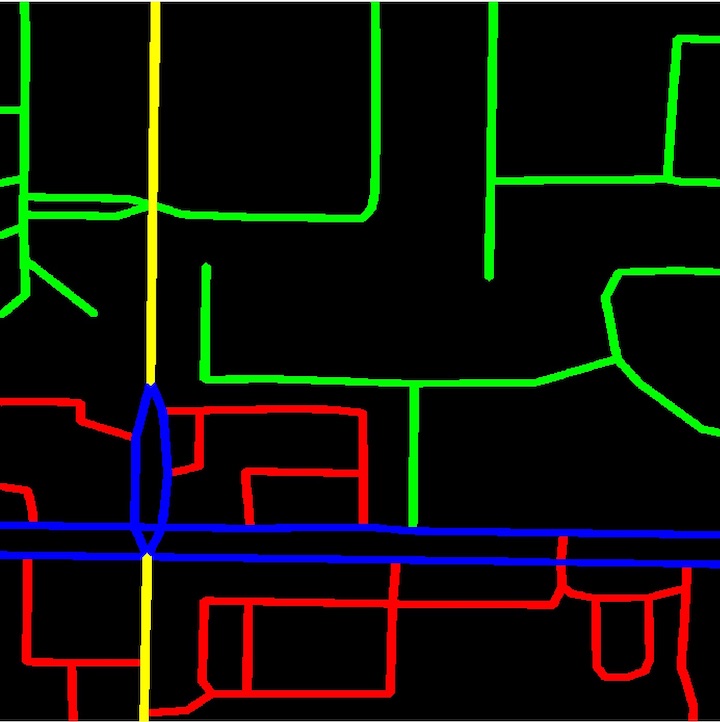}} &
\subfloat [\textbf{Predicted mask}] {\includegraphics[width=0.48\linewidth]{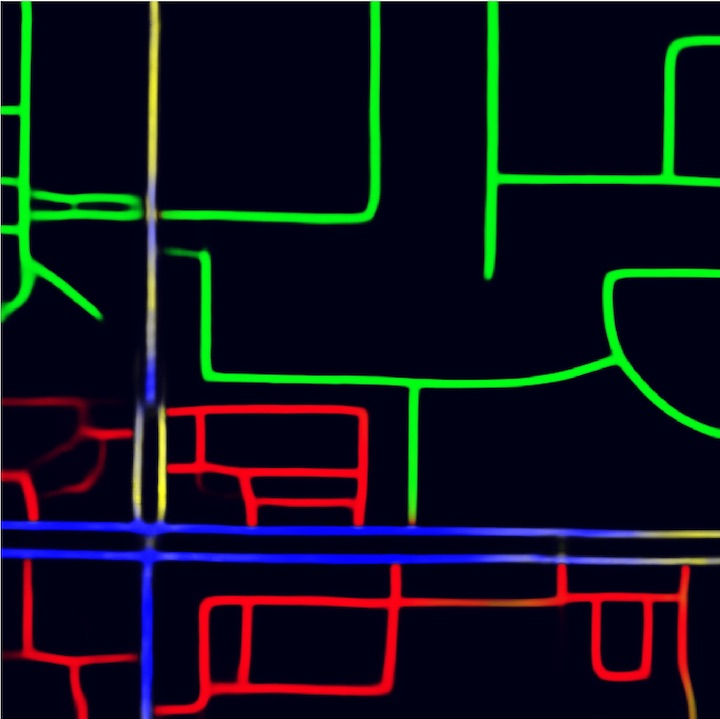}} \\

\subfloat [\textbf{Ground truth network}] {\includegraphics[width=0.48\linewidth]{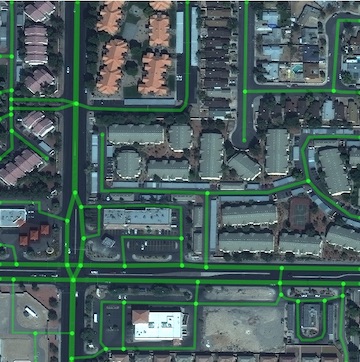}} &
\subfloat [\textbf{Predicted network}] {\includegraphics[width=0.48\linewidth]{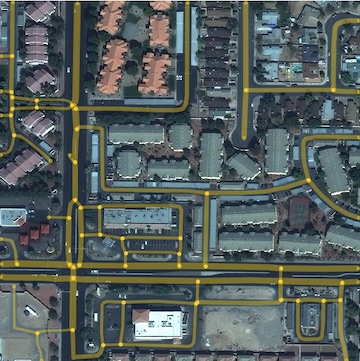}} \\

\end{tabular}

 \caption{\textbf{Algorithm performance on SpaceNet.} (a) Ground truth and (b) predicted multi-class masks: red = 21-30 mph, green = 31-40 mph, blue = 41-50 mph, yellow = 51-60 mph.
(c) Ground truth and (d) predicted graphs overlaid on the SpaceNet test chip; edge widths are proportional to speed limit.  The scores for this proposal are APLS$_{\rm{length}}=0.80$ and APLS$_{\rm{time}}=0.64$. 
% \caption{Ground truth (a) and proposal (b) graphs overlaid on a SpaceNet test chip.  Edge widths are proportional to speed limit.  The scores for this proposals are APLS$_{\rm{length}}=0.80$ and APLS$_{\rm{time}}=0.64$. 
 % Vegas 715
  }
 \label{fig:sn3_chips_comparo}
\end{center}
\vspace{-8pt}
\end{figure}

We compute both APLS and TOPO performance for the $400 \times 400$ meter image chips in the SpaceNet test corpus, utilizing an APLS buffer and TOPO hole size of 4 meters (implying proposal road centerlines must be within 4 meters of ground truth), see Table \ref{tab:f1_snchips}.  An example result is shown in Figure \ref{fig:sn3_chips_comparo}.  
Reported errors ($\pm 1 \sigma$) reflect the relatively high variance of performance among the various test scenes in the four SpaceNet cities.
Table \ref{tab:f1_snchips} indicates that the continuous mask model struggles to accurately reproduce road speeds, due in part to the model's propensity to predict high pixel values for for high confidence regions, thereby skewing speed estimates.  In the remainder of the paper, we only consider the multi-class model.  
Table \ref{tab:f1_snchips} also demonstrates that for the multi-class model the APLS score is still 0.58 when using travel time as the weight, which is only $13\%$ lower than when weighting with geometric distance.  

\begin{table}[h]
  \caption{Performance on SpaceNet Test Chips}
  \vspace{-3pt}
  \label{tab:f1_snchips}
  \small
  \centering
   \begin{tabular}{llll}
%   \begin{tabular}{lccc}
    \toprule
 Model & TOPO & APLS$_{\rm{length}}$ & APLS$_{\rm{time}}$ \\
 \toprule
 % 1 fold
Multi-Class &   $0.53\pm0.23$ &  $0.68\pm0.21$ & $0.58\pm0.21$  \\
%\hline
Continuous &   $0.52\pm0.25$ &  $0.68\pm0.22$ & $0.39\pm0.18$ \\
 % 4 fold
%Multi-Class &   0.54 &  0.64 & 0.56  \\
%Continuous &   0.53 &  0.64 & 0.39 \\
% Model & APLS$_{\rm{length}}$ & APLS$_{\rm{time}}$ & TOPO \\
 %\toprule
%Multi-Class &     0.64 & 0.56 & 0.54 \\
%Continuous &     0.64 & 0.39 & 0.53 \\

    %\toprule
     \bottomrule
  \end{tabular}
   \vspace{-2pt}
\end{table}

%%%%%%%%%%
% only multi-class
\begin{comment}
\begin{table}[h]
  \caption{Performance on SpaceNet Test Chips}
  \label{tab:f1_snchips}
  \small
  \centering
   \begin{tabular}{cccc}
    \toprule
 APLS (length) & APLS (time) & TOPO \\
 \toprule
%(length) & (time) & (length) \\
     0.64 & 0.56 & 0.54 \\
    %\toprule
     \bottomrule
  \end{tabular}
\end{table}
\end{comment}
%%%%%%%%%%

\subsection{Comparison of SpaceNet to OSM}\label{sec:osm}

%OpenStreetMap (OSM) is a great crowd-sourced resource curated by a community of volunteers, and consists primarily of hand-drawn road labels.
% (though other labels exist in certain areas).  
%Though OSM is a great resource, it is incomplete in many areas (see Figure \ref{fig:osm_goof}).  

\begin{figure}[]
\vspace{-5pt}
  \centering
     \includegraphics[width=0.9\linewidth]{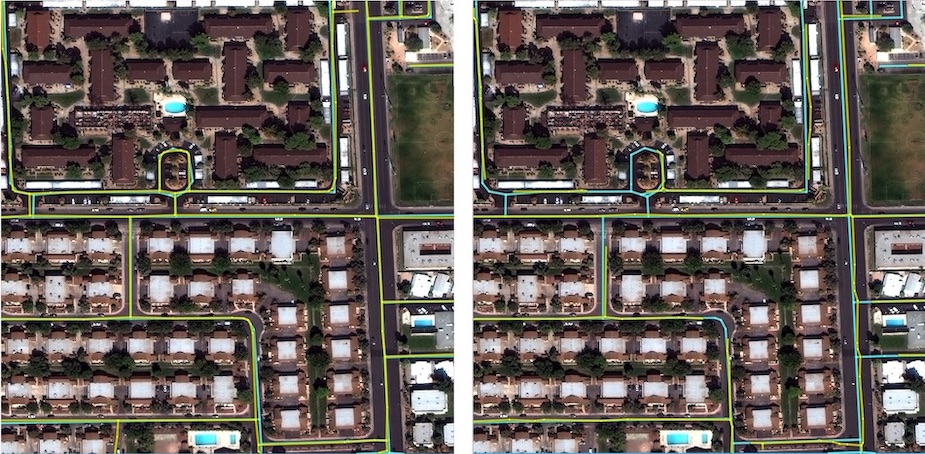}
 \caption{\textbf{SpaceNet compared to OSM.} Road predictions (yellow) and ground truth SpaceNet labels (blue) for a Las Vegas image chip. 
 SpaceNet model predictions (left) score APLS$_{\rm{length}} = 0.94$, 
 while OSM model predictions (right) struggle in this scene with significant offset and missing roads, yielding APLS$_{\rm{length}} = 0.29$.}
 \label{fig:sn_osm_ex} 
\vspace{-8pt}
\end{figure}

As a means of comparison between OSM and SpaceNet labels, we use our algorithm to train two models on SpaceNet imagery. One model uses ground truth segmentation masks rendered from OSM labels, while the other model uses ground truth masks rendered from SpaceNet labels.  Table \ref{tab:osm_vs_sn} displays APLS scores computed over 
a subset of the SpaceNet test chips,
%the same Las Vegas test chips as in Section \ref{sec:ablation}, 
and demonstrates that the model trained and tested on SpaceNet labels is far superior to other combinations, with a $\approx 60 - 100\%$ improvement in APLS$_{\rm{length}}$ score.  This is likely due in part to the the more uniform labeling schema and validation procedures adopted by the SpaceNet labeling team, as well as the superior registration of labels to imagery in SpaceNet data.  The poor performance of the SpaceNet-trained OSM-tested model is likely due to a combination of: different labeling density between the two datasets,
%and OSM label offset when projected onto SpaceNet imagery 
and differing projections of labels onto imagery for SpaceNet and OSM data.  Figure \ref{fig:sn_osm_ex} and Appendix C illustrate the difference between predictions returned by the OSM and SpaceNet models.

\begin{table}[h]
  \caption{OSM and SpaceNet Performance}
  \vspace{-3pt}
  \label{tab:osm_vs_sn}
  \small
  \centering
   \begin{tabular}{llllll}
    \toprule
     Training Labels & Test Labels & APLS$_{\rm{length}}$ \\
    \toprule
    OSM & OSM & 0.47 \\
    OSM & SpaceNet & 0.46 \\
    SpaceNet & OSM & 0.39 \\
    SpaceNet & SpaceNet & 0.77 \\
     \bottomrule
  \end{tabular}
  \vspace{-5pt}
\end{table}

%%%%%%%%%%%%%%%%%
\begin{comment}
\begin{figure}[]
\vspace{-5pt}
\begin{center}
\setlength{\tabcolsep}{0.3em}
\begin{tabular}{cc}
\vspace{-5pt}

\subfloat [\textbf{OSM}] {\includegraphics[width=0.48\linewidth]{1019_osm_zoom.jpg}} &
\subfloat [\textbf{SpaceNet}] {\includegraphics[width=0.48\linewidth]{1019_spacenet_zoom.jpg}} \\

\end{tabular}

 \caption{\textbf{SpaceNet compared to OSM.} Road predictions (yellow) and ground truth SpaceNet labels (blue) for a sample Las Vegas image chip. OSM model predictions (a) are slightly more offset from ground truth labels than SpaceNet model predictions (b).} 
 \label{fig:sn_osm_ex} 
\end{center}
\vspace{-10pt}
\end{figure}
\end{comment}
%%%%%%%%%%%%%%%%%

%%%%%%%%%%%%%%%%%%%%%%%%%%
\subsection{Ablation Study}\label{sec:ablation}

In order to assess the relative importance of various improvements to our baseline algorithm, we perform ablation studies on the final algorithm.  For evaluation purposes we utilize the 
the same subset of test chips as in Section \ref{sec:osm}, 
%Las Vegas SpaceNet test data 
and the APLS$_{\rm{length}}$ metric.  %Performance is detailed in Table \ref{tab:ablation}. 
Table \ref{tab:ablation} demonstrates that advanced post-processing significantly improves scores.  Using a more complex architecture also improves the final prediction.  Applying four folds improves scores very slightly, though at the cost of significantly increased algorithm runtime. Given the minimal improvement afforded by the ensemble step, all reported results use only a single model.

\begin{table}[]
  \caption{Road Network Ablation Study}
   \vspace{-3pt}
 \label{tab:ablation}
  \small
  \centering
   \begin{tabular}{llr}
    \toprule
     & Description & APLS \\
    \toprule
    1 & Extract graph directly from simple U-Net model & 0.56 \\
    2 & Apply opening, closing, smoothing processes & 0.66 \\
    3 & Close larger gaps using edge direction and length & 0.72 \\
    %4 & Use 4 folds and ResNet34 + U-Net architecture & 0.77 \\
    4 & Use ResNet34 + U-Net architecture & 0.77 \\
    5 & Use 4 fold ensemble  & 0.78 \\
    %5 & Use 4 folds with ResNet34 + U-Net architecture & 0.78 \\
    \bottomrule
  \end{tabular}
  \vspace{-8pt}
\end{table}

\subsection{Large Area SpaceNet Results}\label{sec:CRESI_results}
%\section{CRESI Results}\label{sec:CRESI_results}

\begin{figure}[]
\vspace{-5pt}
\begin{center}
\centering
\setlength{\tabcolsep}{0.3em}
\begin{tabular}{cc}
\vspace{-5pt}
\subfloat {\includegraphics[width=0.96\linewidth]{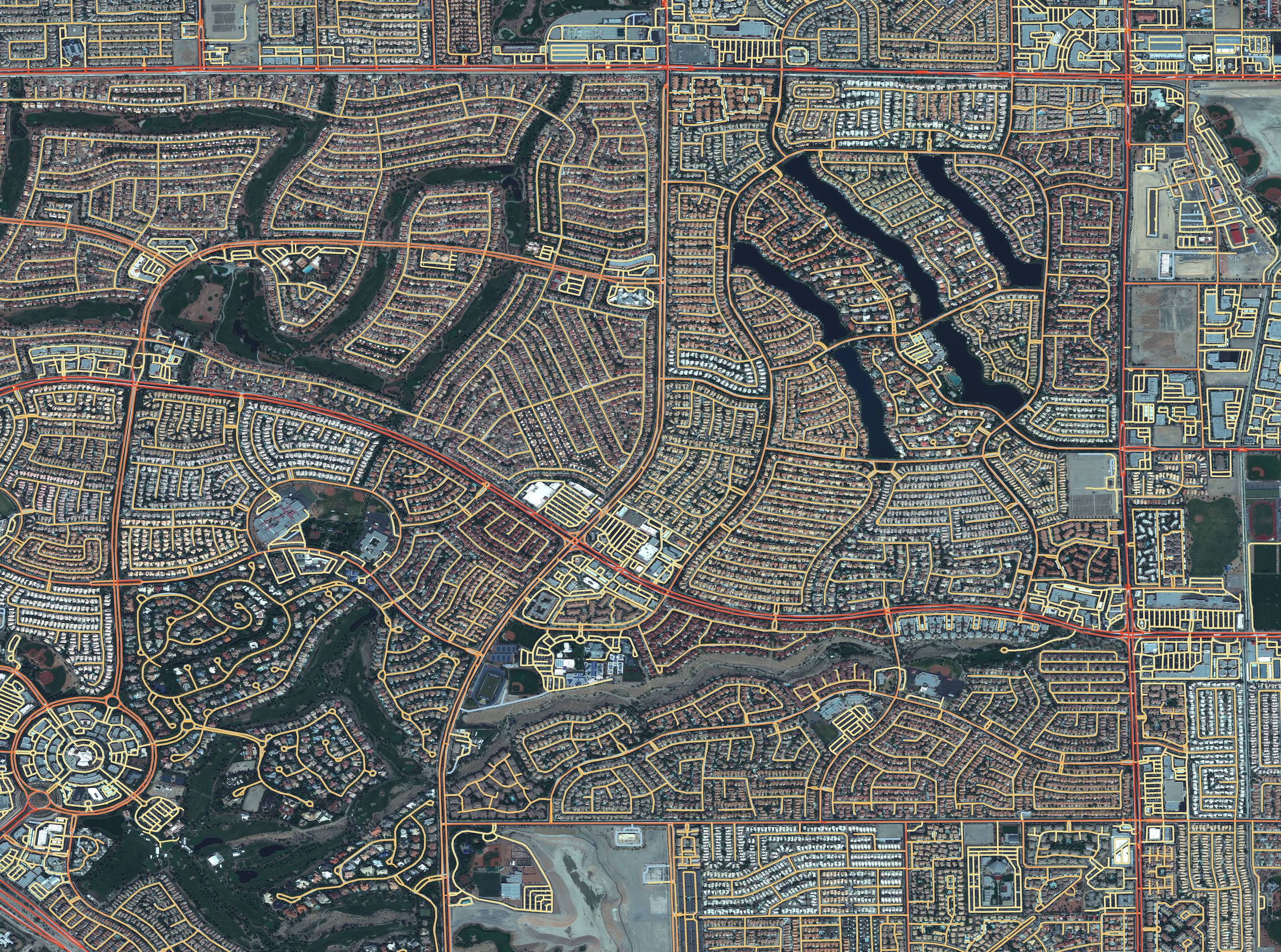}}\\
\subfloat {\includegraphics[width=0.96\linewidth]{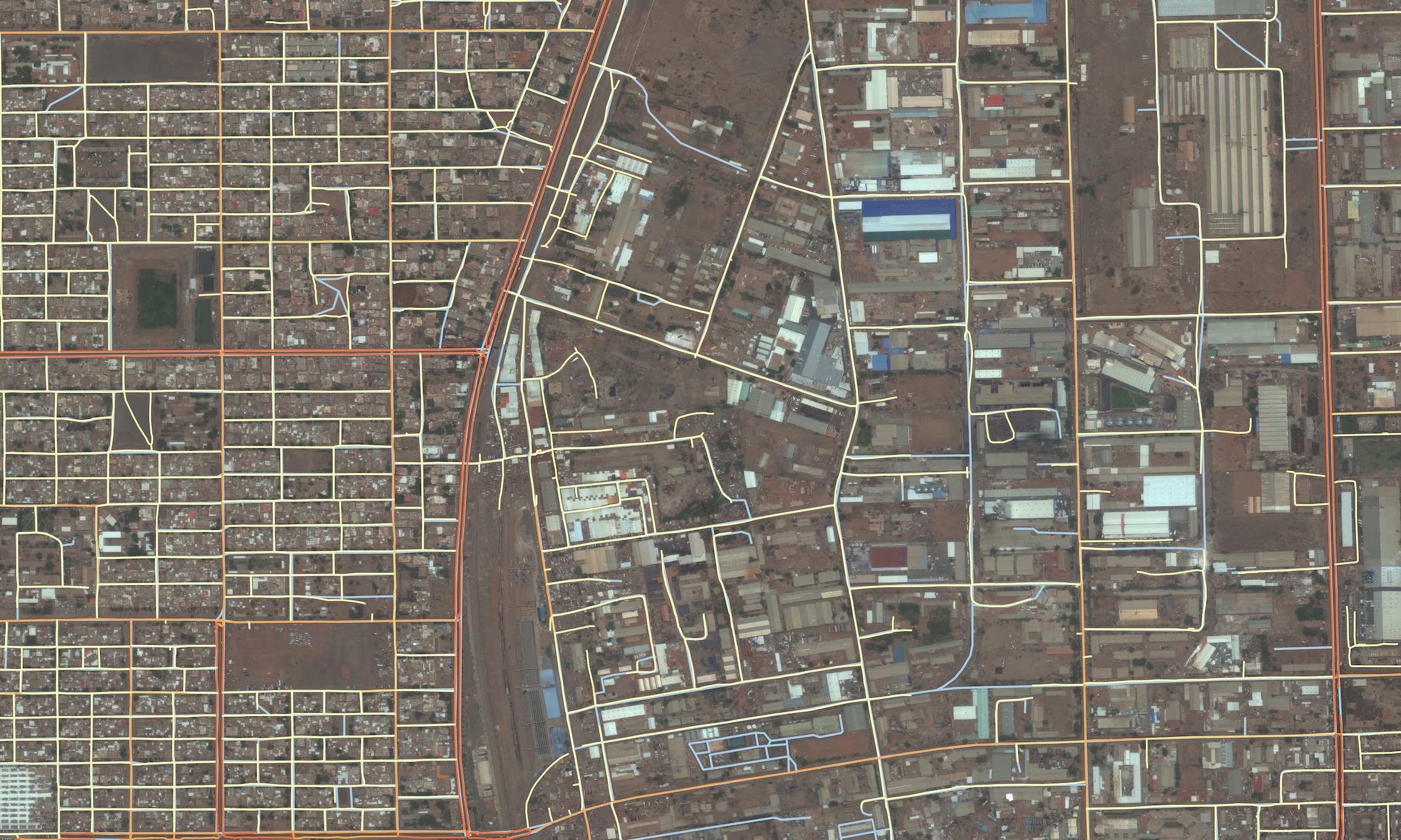}}\\
\end{tabular}
\vspace{-3pt}
\caption{\textbf{CRESIv2 road speed.} Output of CRESIv2 inference as applied to the SpaceNet large area test dataset.  
Predicted roads are colored by inferred speed limit, from yellow (20 mph) to red (65 mph).  Ground truth labels are shown in gray.
{Top:} Las Vegas: APLS$_{\rm{length}} = 0.85$ and APLS$_{\rm{time}} = 0.82$.
{Bottom:} A smaller region of Khartoum: APLS$_{\rm{length}} = 0.71$ and the APLS$_{\rm{time}} = 0.67$.
}
\label{fig:res_vegas0_khartoum2}
\end{center}
\vspace{-15pt}
\end{figure}

We apply the CRESIv2 algorithm described in Table \ref{tab:algo2} to the 
large area SpaceNet test set covering 608 km$^2$.
%test images of Section \ref{sec:testdata}. 
Evaluation takes place with the APLS metric adapted for large images (no midpoints along edges and a maximum of 500 random control nodes), along with the TOPO metric, using a buffer size (for APLS) or hole size (for TOPO) of 4 meters.  
%Previous road network detection studies have typically used a TOPO hole size of $> 10$ meters (citiations?).   
%We report scores in Table \ref{tab:test_perf} as the weighted (by road length) mean and standard deviation of the test regions of Table \ref{tab:test_regs}.  
We report scores in Table \ref{tab:test_perf} as the mean and standard deviation of the test regions of in each city.  
Table \ref{tab:test_perf} reveals an overall $\approx4\%$ decrease in APLS score when using speed versus length as edge weights.  
This is somewhat less than the decrease of 13\% noted in Table \ref{tab:f1_snchips}, due primarily to the fewer edge effects from larger testing regions.
Table \ref{tab:test_perf} indicates a large variance in scores across cities; locales like Las Vegas with wide paved roads and sidewalks to frame the roads are much easier than Khartoum, which has a multitude of narrow, dirt roads and little differentiation in color between roadway and background.
%The total APLS$_{\rm{time}}$ score of 0.67.
%Figure \ref{fig:tricky} illustrates that road network extraction is possible even for atypical lighting conditions and off-nadir observation angles (Paris), and also that road extraction is possible (though imperfect) in locales with both a multitude of dirt roads and low contrast between roads and background (Khartoum).  
%CRESI lends itself to optimal routing in complex road systems.
Figure \ref{fig:res_vegas0_khartoum2} and Appendix D
%Figures \ref{fig:res_vegas0} (Las Vegas) and \ref{fig:res_khartoum2} (Khartoum)
 display the graph output for various urban environments.  
%, while Figure \ref{fig:res_shang0} zooms in on an industrial portion of Shanghai.

%Since the algorithm output is a \rm{ networkx} graph structure, myriad graph algorithms can be easily applied.  In addition, since we retain geographic information throughout the graph creation process, we can overlay the graph nodes and edges on the original GeoTIFF that we input into our model.  
%Figures \ref{fig:res_r00} and \ref{fig:res_r1} display portion of Las Vegas and Paris, respectively, overlaid with the inferred road network.  

%Figure \ref{fig:res_r1}demonstrates that road network extraction is possible even for atypical lighting conditions and off-nadir observation angles, and also that CRESI lends itself to optimal routing in complex road systems.
% even in atypical lighting conditions.  
	%s possible with the inferred network by plotting the shortest path between nodes of interest in the graph. 

%%%%%%%%%
%%%%%%%%%

\begin{table}[h]
% see CRESI_supp for all data
  \caption{SpaceNet Large Area Performance}
  \vspace{-3pt}
  \label{tab:test_perf}
  \small
  \centering
   \begin{tabular}{llll}
    \toprule
     Test Region & TOPO & APLS$_{\rm{length}}$ & APLS$_{\rm{time}}$  \\
% 1 fold results
   \toprule
    	Khartoum  	& $0.53 \pm 0.09$  & $ 0.64 \pm 0.10$ 	& $0.61 \pm 0.05$ \\
    	Las Vegas 	& $0.63 \pm 0.02$  & $ 0.81 \pm 0.04$ 	& $0.79 \pm 0.02$ \\
    	Paris		 	& $0.43 \pm 0.01$  & $ 0.66 \pm 0.04$ 	& $0.65 \pm 0.02$ \\
    	Shanghai  	& $0.45 \pm 0.03$  & $ 0.55 \pm 0.13$ 	& $0.51 \pm 0.11$ \\
	\hline
	Total 		 & $0.51 \pm 0.02$ & $ 0.67 \pm 0.04$    & $0.64 \pm 0.03$ \\

%    	Khartoum  	& $0.53 \pm 0.09$  & $ 0.61 \pm 0.07$ 	& $0.60 \pm 0.06$ \\
 %   	Las Vegas 	& $0.63 \pm 0.02$  & $ 0.80 \pm 0.02$ 	& $0.76 \pm 0.01$ \\
 %   	Paris		 	& $0.43 \pm 0.01$  & $ 0.58 \pm 0.02$ 	& $0.56 \pm 0.02$ \\
 %   	Shanghai  	& $0.45 \pm 0.03$  & $ 0.50 \pm 0.10$ 	& $0.46 \pm 0.09$ \\
%	\hline
%	Total 		 & $0.56 \pm 0.09$ & $ 0.69 \pm 0.15$    & $0.65 \pm 0.14$ \\

 %% 4 fold results
 %   \toprule
%    	Khartoum  	& $0.54 \pm 0.08$  & $ 0.63 \pm 0.08$ 	& $0.59 \pm 0.06$ \\
 %  	Las Vegas 	& $0.65 \pm 0.02$  & $ 0.82 \pm 0.02$ 	& $0.78 \pm 0.02$ \\
 %   	Paris		 	& $0.44 \pm 0.03$  & $ 0.60 \pm 0.03$ 	& $0.60 \pm 0.02$ \\
 %   	Shanghai  	& $0.45 \pm 0.03$  & $ 0.51 \pm 0.01$ 	& $0.47 \pm 0.08$ \\
%	\hline
%	Total 		 & $0.58 \pm 0.09$ & $ 0.71 \pm 0.15$    & $0.67 \pm 0.15$ \\
    \bottomrule
  \end{tabular}
  \vspace{-1pt}
\end{table}

\begin{comment}
\begin{figure}[]
\vspace{-6pt}
\begin{center}
\setlength{\tabcolsep}{0.3em}
\begin{tabular}{c}
\vspace{-6pt}

\subfloat {\includegraphics[width=0.97\linewidth]{khartoum_2_edges.jpg}} \\
\vspace{-5pt}
\subfloat {\includegraphics[width=0.97\linewidth]{paris_0_edges.jpg}} \\
%\subfloat [\textbf{Khartoum}] {\includegraphics[width=0.97\linewidth]{khartoum_2_edges.jpg}} \\
%\subfloat [\textbf{Paris}] {\includegraphics[width=0.97\linewidth]{paris_0_edges.jpg}} \\

\end{tabular}

 \caption{\textbf{CRESIv2 outputs.} Top: Road predictions (yellow) overlaid on a portion of the Khartoum$\_2$ test region with a high percentage of dirt roads.  Bottom: Road predictions overlaid on a portion of the Paris$\_0$ test region in atypical (dark) lighting conditions and $19^{\circ}$ off-nadir.} 
 \label{fig:tricky} 
\end{center}
\vspace{-5pt}
\end{figure}
\end{comment}

%%%%%%%%%
\begin{comment}
%\begin{figure*}[h]
\begin{figure}%[h]
\begin{center}
\includegraphics[width=0.97\linewidth]{shanghai_0_shp.jpg}
\end{center}
\caption{Output of CRESIv2 inference as applied to the Shanghai$\_0$ test region. 
%There are 883 km of  roads in this region.
}
\label{fig:res_r0}
\end{figure}
\end{comment}
%%%%%%%%%

% Vegas 0 and Khartoum 2

\subsection{Google / OSM Results}

Applying our methodology to 60 cm Google imagery with OSM labels achieves state of the art results.  
% Our method achieves reasonable results on 60 cm Google imagery with OSM labels as well.
%In Table \ref{tab:rt_cities} we report scores for the same 4m APLS buffer and TOPO hole size used above.  
For the same 4 m APLS buffer 
%and TOPO hole size 
used above, we achieve a score of
APLS$_{\rm{length}} = 0.53 \pm 0.11$.
%and TOPO = $0.13 \pm 0.04$.
This score is consistent with the results of Table \ref{tab:osm_vs_sn}, and compares favorably to previous methods (see Table \ref{tab:comparo} and Figure \ref{fig:nyc}).  
% Figure 
% \ref{fig:boston} 
% \ref{fig:nyc} illustrates that our method  is far more complete and misses fewer small roadways and intersections than previous methods.  
%See the following Section for further analysis.

%%%%%%%%%%%%%
\begin{comment}
\begin{figure}
%\vspace{-5pt}
  \centering
     \includegraphics[width=0.95\linewidth]{boston_ox_plot.jpg}
       \caption{\textbf{Inference on 60 cm imagery.} Prediction for the Boston test region of the Google / OSM dataset.
       The APLS$_{\rm{length}}$  score for this region is 0.53.  }
  \label{fig:boston}
\vspace{-6pt}
\end{figure}
\end{comment}
%%%%%%%%%%%%%%%

\subsection{Comparison to Previous Work}

\begin{figure}[]
\vspace{-5pt}
\begin{center}
\setlength{\tabcolsep}{0.3em}
\begin{tabular}{cc}
\vspace{-5pt}

\captionsetup[subfigure]{labelformat=empty}

\subfloat {\includegraphics[width=0.49\linewidth]{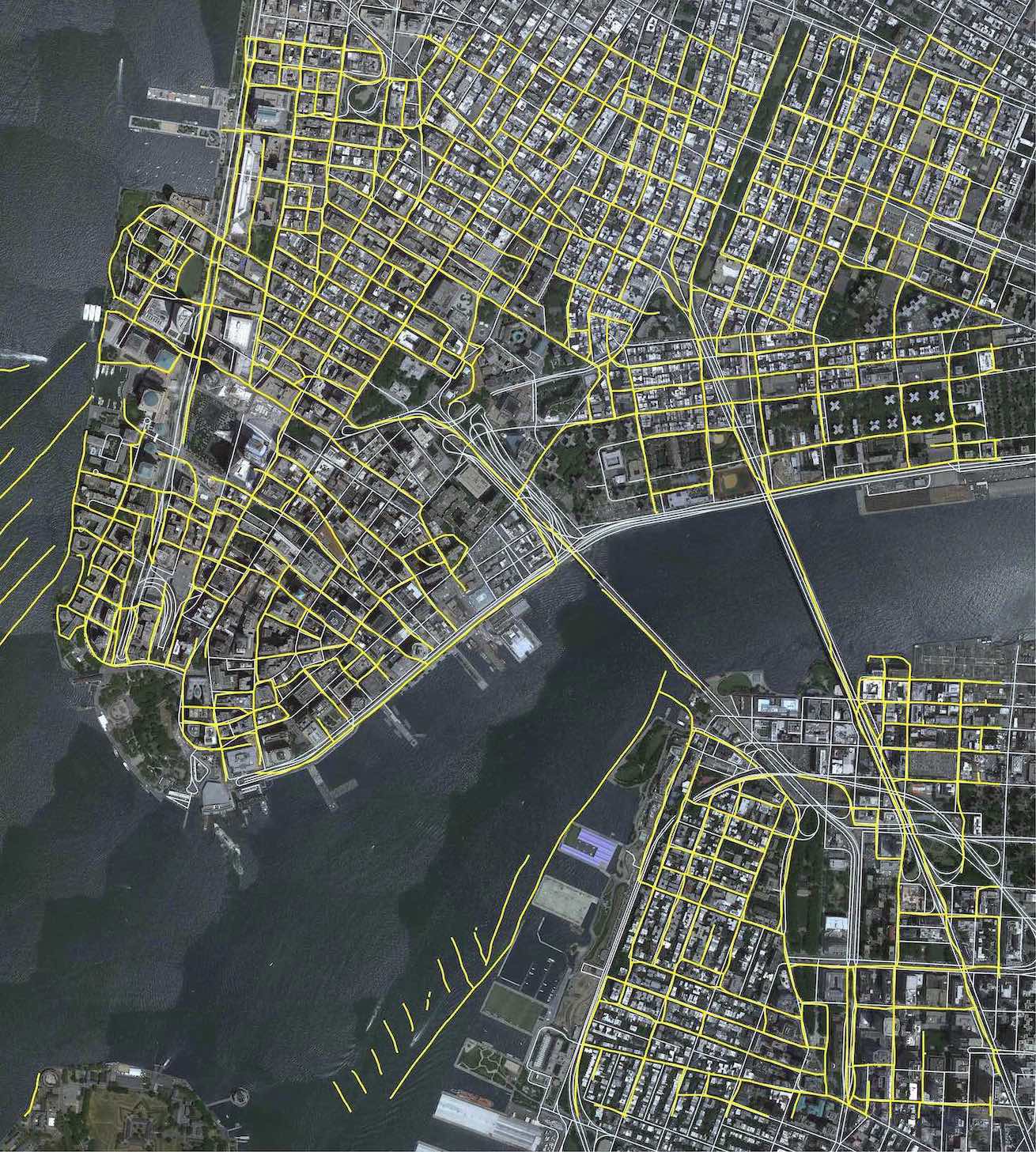}} &
\subfloat {\includegraphics[width=0.49\linewidth]{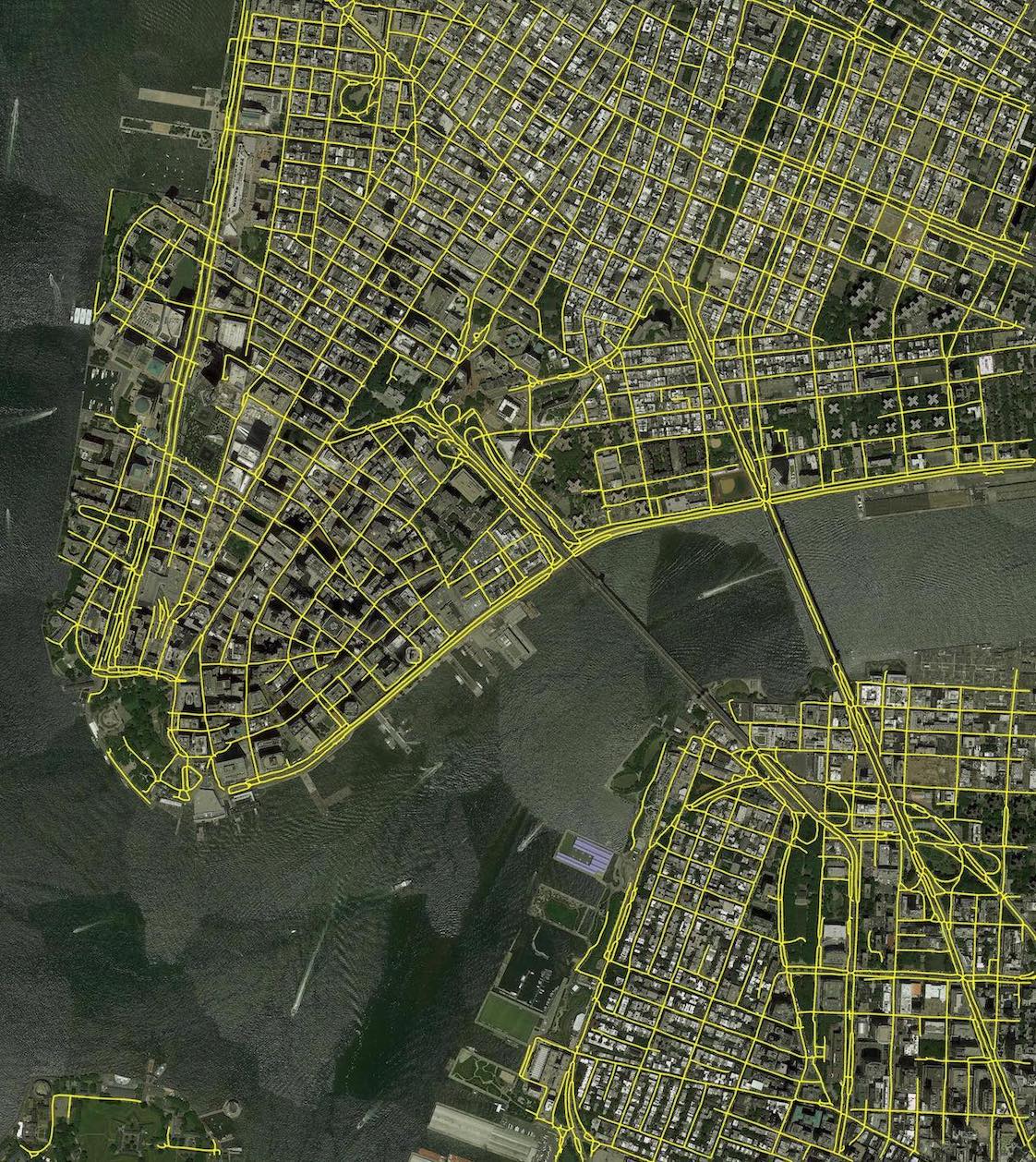}} \\
\subfloat  [\textbf{RoadTracer}] {\includegraphics[width=0.49\linewidth]{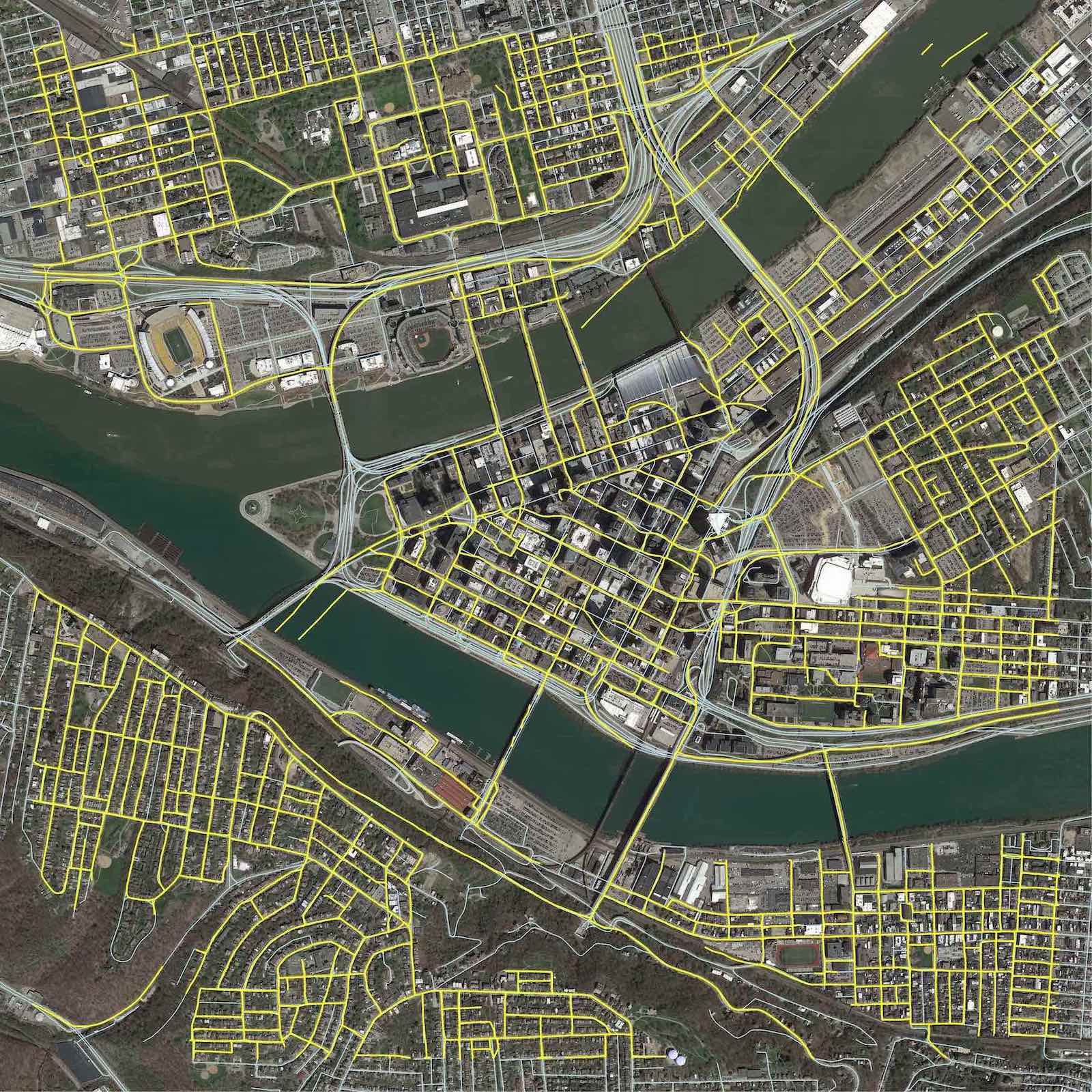}} &
\subfloat  [\textbf{CRESIv2}] {\includegraphics[width=0.49\linewidth]{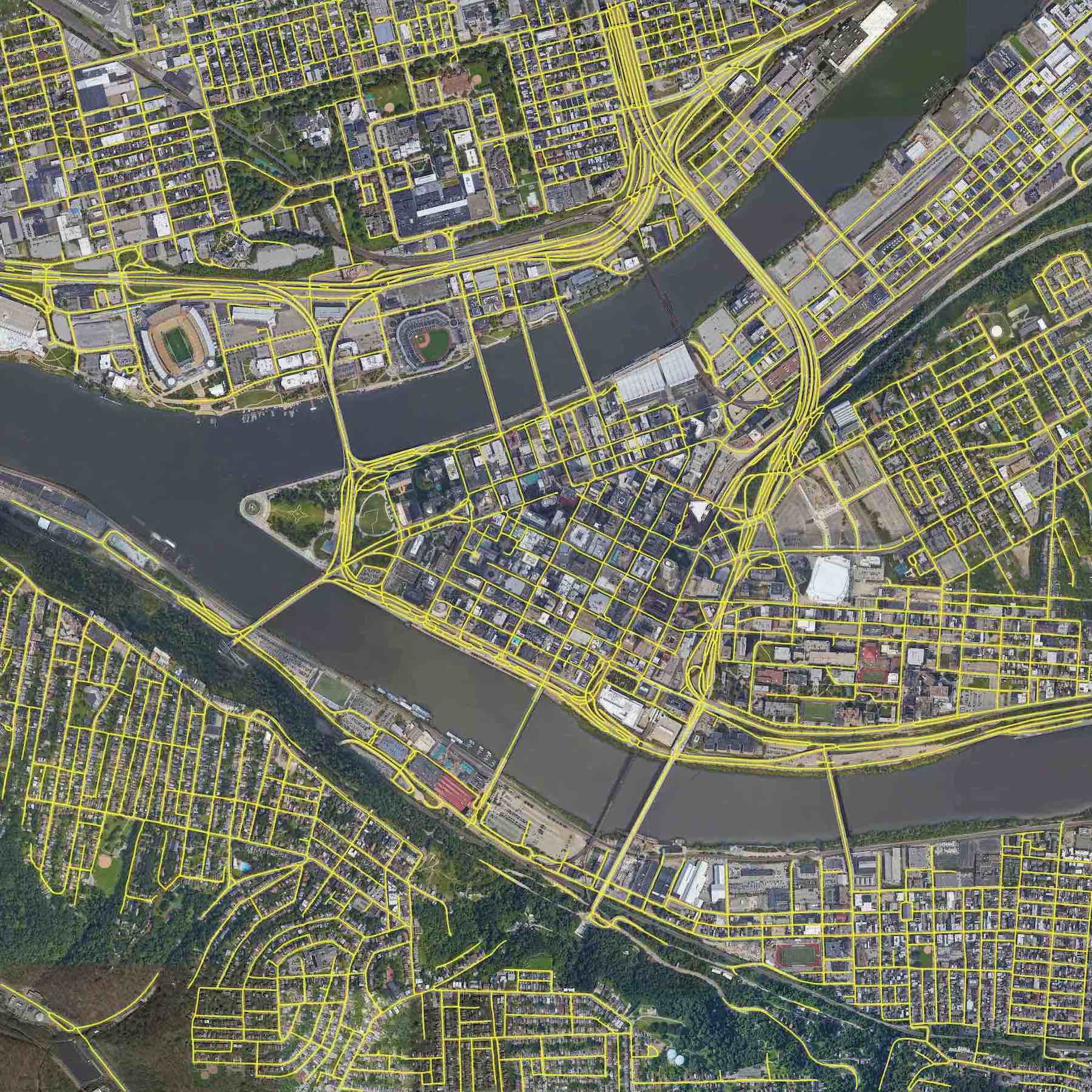}} \\

%\subfloat [\textbf{RoadTracer - New York}] {\includegraphics[width=0.48\linewidth]{new_york_roadtracer_highres.jpg}} &
%\subfloat [\textbf{CRESIv2 - New York}] {\includegraphics[width=0.48\linewidth]{new_york_ox_plot_clip.jpg}} \\
%\subfloat [\textbf{RoadTracer - Pittsburgh}] {\includegraphics[width=0.48\linewidth]{pittsburgh_roadtracer_highres.jpg}} &
%\subfloat [\textbf{CRESIv2 - Pittsburgh}] {\includegraphics[width=0.48\linewidth]{pittsburgh_ox_plot.jpg}} \\

\end{tabular}

 \caption{\textbf{New York City (top) and Pittsburgh (bottom) Performance.} (Left) RoadTracer prediction \cite{roadtracer_ims}. (Right) Our CRESIv2 prediction over the same area. 
% \caption{\textbf{New York City / Pittsburgh Performance.} (a/c) RoadTracer prediction \cite{roadtracer_ims} (b/d) Our CRESIv2 prediction over the same areas. 
  }
  
 \label{fig:nyc}
\end{center}
\vspace{-15pt}
\end{figure}

Table \ref{tab:comparo} demonstrates that CRESIv2 improves upon existing methods for road extraction, 
both on the $400 \times 400$ m SpaceNet image chips at 30 cm resolution, as well as 60 cm Google satellite imagery with OSM labels.
To allow a direct comparison in Table \ref{tab:comparo}, we report TOPO scores with the 15 m hole size used in \cite{roadtracer}.  
A qualitative comparison is shown in Figure \ref{fig:nyc} and Appendices E and F,
illustrating that our method  is more complete and misses fewer small roadways and intersections than previous methods.

\begin{table}[h]
% see CRESI_supp for all data
  \caption{Performance Comparison}
  \vspace{-3pt}
  \label{tab:comparo}
  \small
  \centering
   \begin{tabular}{lcc}
    \toprule
     %Test Region & TOPO & APLS$_{\rm{length}}$ & APLS$_{\rm{time}}$  \\
     Algorithm	 	 	& 	Google / OSM  		& SpaceNet  \\
    					&      (TOPO)		& (APLS$_{\rm{length}}$) \\
    \toprule     
    DeepRoadMapper \cite{deeproadmapper}	&	0.37				& 	0.51\footnotemark[1] \\
    RoadTracer \cite{roadtracer}	 	 	& 	0.43 				&      0.58\footnotemark[1] \\
    OrientationLearning \cite{Batra_2019_CVPR}	&	   -	 			&	0.64 \\
    CRESIv2 (Ours) 		& 	\bf{0.53}			& 	\bf{0.67} \\    
    
   \bottomrule
  \end{tabular}
   \begin{tablenotes}
     \item[1] $^1$ from Table 4 of \cite{Batra_2019_CVPR}
    % \item[2] tablefootnote 2
   \end{tablenotes}  
  
 % \caption{* from \cite{Batra_2019_CVPR}}
%  \vspace{-5pt}
\footnotetext[1]{table footnote 1}

\end{table}

\captionsetup[subfigure]{labelformat=empty}

\section{Discussion}\label{sec:discussion}

CRESIv2 improves upon previous methods in extracting road topology from satellite imagery, 
%with a 5\% improvement over previous efforts on the SpaceNet dataset, and a 23\% improvement over previous efforts with Google satellite imagery + OSM labels.  
%Given the difference in approaches, 
The reasons for our 5\% improvement over the Orientation Learning method applied to SpaceNet data
are difficult to pinpoint exactly, but our custom dice + focal loss function (vs the SoftIOU loss of \cite{Batra_2019_CVPR}) is a key difference.  The enhanced ability of CRESIv2 to disentangle areas of dense road networks accounts for most of the 23\% improvement over the RoadTracer method applied to Google satellite imagery + OSM labels.  

We also introduce the ability to extract route speeds and travel times.  Routing based on time shows only a $3-13\%$ decrease from distance-based routing,
indicating that true optimized routing is possible with this approach.  The aggregate score of APLS$_{\rm{time}} = 0.64$ implies that travel time estimates will be within $\approx \frac{1}{3}$ of the ground truth.
% across varied locales.   

\begin{figure}%[h]
\vspace{-3pt}
  \centering
\includegraphics[width=0.77\linewidth]{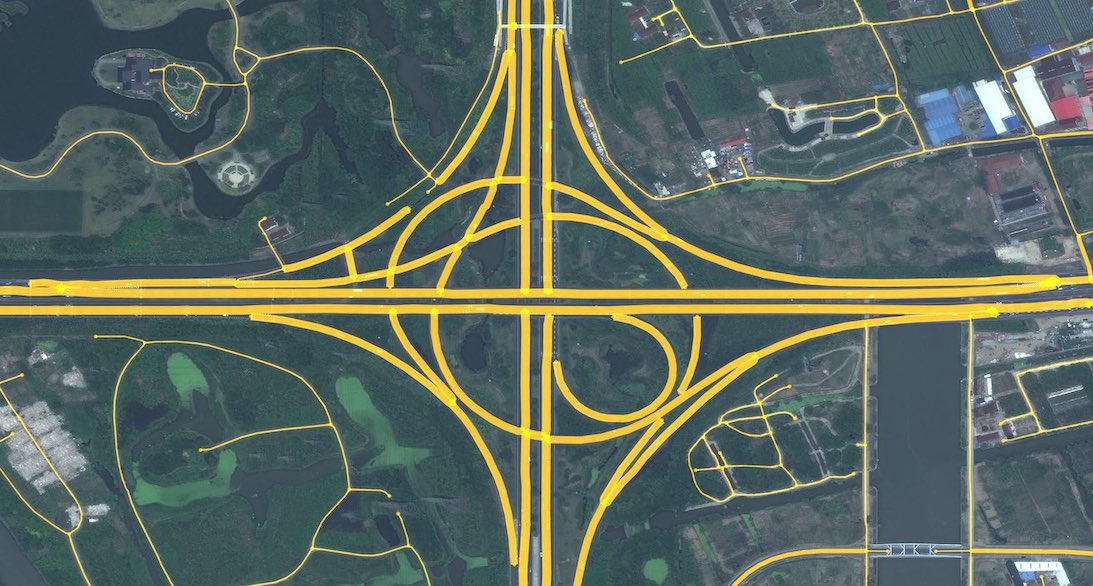}
\caption{\textbf{CRESIv2 challenges}. 
While the pixel-based score of this Shanghai prediction is high, correctly connecting roadways in complex intersections remains elusive.
%Zoom of CRESIv2 inference over an intersection in Shanghai.  While the pixel-base score of this prediction is very high, correctly connecting roadways in complex intersections remains elusive.
%Zoom of CRESIv2 inference over an intersection in Shanghai.  While the pixel-base score of this prediction is very high, correctly connecting roadways in complex intersections remains elusive.
}
\label{fig:res_shang0}
\vspace{-8pt}
\end{figure}

As with most approaches to road vector extraction, complex intersections are a challenge with CRESIv2.  
While we attempt to connect gaps based on road heading and proximity, overpasses and onramps remain difficult (Figure \ref{fig:res_shang0}).

% Shanghai 0

CRESIv2 has not been fully optimized for speed, but even so inference runs at a rate of $280 \,{\rm km}^2 / \, {\rm hour}$  on a machine with a single Titan X GPU.  
At this speed, a 4-GPU cluster could map the entire $9100 \, {\rm km}^2$ area of Puerto Rico in $\approx 8$ hours, a significant improvement over the two months required by human labelers  \cite{osm_maria}.

%Inference code has not been optimized for speed, but even so inference runs at a rate of $170 \,{\rm km}^2$  (the approximate area of Washington D.C.) $/ \, {\rm hour}$ %(approximately the area of Washington D.C.) 
%on a single GPU machine.  %On a four GPU cluster the speed is a minimum of $370 \, {\rm km}^2 / \, {\rm hour}$.  

\section{Conclusion}\label{sec:conclusion}

Optimized routing is crucial to a number of challenges, from humanitarian to military. Satellite imagery may aid greatly in determining efficient routes, particularly in cases involving natural disasters or other dynamic events where the high revisit rate of satellites may be able to provide updates far more quickly than terrestrial methods. 

%In this paper we demonstrated methods to extract city-scale road networks directly from remote sensing imagery.  
%GPU memory limitations constrain segmentation algorithms to inspect images of size $\sim2000 \times 2000$ pixels in extent, yet any eventual application of road inference must be able to process images far larger than a mere $\sim\frac{1}{3} \rm{km}^2$. Accordingly, we demonstrate methods to infer road networks and travel times for input images of arbitrary size. 
In this paper we demonstrated methods to extract city-scale road networks directly from remote sensing images of arbitrary size, regardless of GPU memory constraints.  
This is accomplished via a multi-step algorithm that segments small image chips, extracts a graph skeleton, refines nodes and edges, stitches chipped predictions together, extracts the underlying road network graph structure, and infers speed limit / travel time properties for each roadway.  
Our code is publicly available at \texttt{github.com/CosmiQ/cresi}.  

%Measuring performance with the APLS graph theoretic metric we observe superior performance for models trained and tested on SpaceNet data over OSM data.  
Applied to SpaceNet data, we observe a 5\% improvement over published methods, and when using OSM data our method provides a significant (+23\%) improvement over existing methods.  
Over a diverse test 
set that includes atypical lighting conditions, off-nadir observation angles, and locales with a multitude of dirt roads,
%a large test area covering four SpaceNet test cities, 
we achieve a total score of 
APLS$_{\rm length}= 0.67$, and nearly equivalent performance when optimizing for travel time: APLS$_{\rm time} = 0.64$.
Inference speed is a brisk $\geq280 \, {\rm km} ^2 \, / \, \rm{hour} / \rm{GPU}$.
%Prediction is reasonably quick, with an inference speed of $\geq280 \, {\rm km} ^2 \, / \, \rm{hour}$ on a single GPU.
%APLS$_{\rm length}= 0.71 \pm 0.15$, 
%and APLS$_{\rm time} = 0.67 \pm 0.15$, 
%with an inference speed of $\geq170 \, {\rm km} ^2 \, / \, \rm{hour}$.  

While automated road network extraction is by no means a solved problem, the CRESIv2 algorithm 
%improves upon and extends existing methods, 
demonstrates that true time-optimized routing is possible,
with potential benefits to applications such as disaster response where rapid map updates are critical to success.  

\vspace{5pt}
\begin{footnotesize}
\noindent
{\bf Acknowledgments}
We thank Nick Weir, Jake Shermeyer, and Ryan Lewis for their insight, assistance, and feedback.
\end{footnotesize}

%----------------------------
{\small
\bibliographystyle{ieee}
\bibliography{bib}
}

%%%%%%%%%%%%%%%%%%%%%
% Supplemental
% No numbering for Appendices
\setcounter{secnumdepth}{0}

\newpage
\clearpage
\newpage

%%%%%%%%%%%%%%%%%%%%%%%%%%
\section*{Appendix A. Road Speed Assignment}

See \cite{spacenet} for details on the precise labeling guidelines and road type definitions. We utilize road type (motorway, primary, secondary, tertiary, residential, unclassified, cart track) and road surface type (paved, non-paved) to assign speed to each edge.  

Speed is assigned with Table \ref{tab:speed}, using the Oregon guidelines for road speed \cite{oregon_speed}.

\begin{table}[h]
  \caption{Road Speeds (mph)}
  \label{tab:speed}
  \small
  \centering
   \begin{tabular}{llllll}
    \toprule
    Road Type & 1 Lane & 2 Lane & 3+ Lane \\ %& 4 Lane \\
    \toprule
    Motorway    & 55 & 55 & 65 \\ % & 65 \\
    Primary       & 45 & 45 & 55 \\ %&55 \\
    Secondary  & 35 & 35 & 45 \\ %& 45 \\
    Tertiary        & 30 & 30 & 35 \\ %& 35 \\
    Residential  & 25 & 25 & 30 \\ %& 30 \\
    Unclassified & 20 & 20 & 20 \\ %& 20 \\
    Cart Track    & 20 & 20 & 20 \\ %& 20 \\
\bottomrule
  \end{tabular}
\end{table}

For each non-paved roadway, the speed from Table \ref{tab:speed} is multiplied by 0.75 to give the final speed.

\newpage
%\clearpage

\section*{Appendix B. Large Area Test Data}

\begin{figure}[h]
\begin{center}
\includegraphics[width=0.85\linewidth]{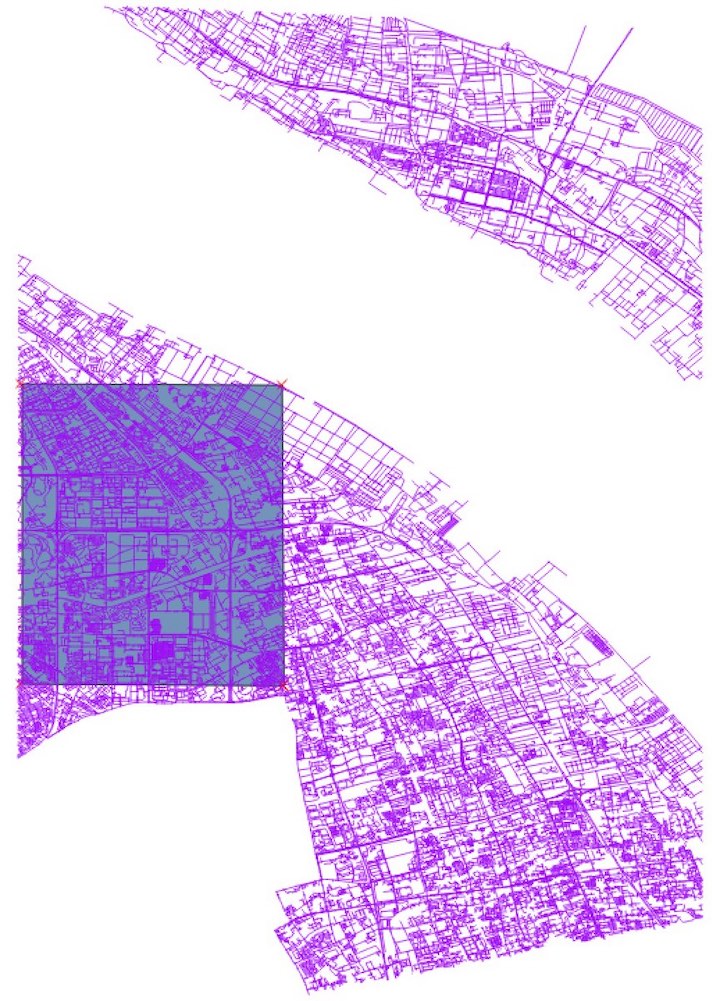}
\end{center}
\caption{SpaceNet road vector labels over Shanghai (purple).  The label boundary is discontinuous and irregularly shaped, so we define rectangular regions for testing purposes (e.g.~ the blue region denotes test region Shanghai$\_$0).}
\label{fig:shanghai_reg0}
\end{figure}

Details of the large are testing regions for SpaceNet data are shown in Table  \ref{tab:test_regs}, and an example test area displayed in Figure \ref{fig:shanghai_reg0}.

\begin{table}[h]
  \caption{Test Regions}
  \vspace{-3pt}
  \label{tab:test_regs}
  \small
  \footnotesize
  \centering
   \begin{tabular}{lll}
    \toprule
     Test Region & Area & Road Length \\
     & (Km$^2$) & (Total Km) \\
    \toprule
    	Khartoum$\_0$ 	& 3.0 	& 76.7 \\
	Khartoum$\_1$ 	& 8.0 	& 172.6 \\
	Khartoum$\_2$ 	& 8.3 	& 128.9\\
	Khartoum$\_3$ 	& 9.0 	& 144.4 \\
	Las$\_$Vegas$\_0$ 	& 68.1	& 1023.9\\
	Las$\_$Vegas$\_1$	& 177.0	& 2832.8\\
	Las$\_$Vegas$\_2$	& 106.7 	& 1612.1\\
	Paris$\_0$ 		& 15.8	& 179.9\\
	Paris$\_1$		& 7.5		& 65.4\\
	Paris$\_2$		& 2.2		& 25.9\\
	Shanghai$\_0$		& 54.6	& 922.1\\
	Shanghai$\_1$		& 89.8	& 1216.4\\
	Shanghai$\_2$		& 87.5	& 663.7\\
	\hline
	Total			& 608.0	& 9064.8\\
    \bottomrule
  \end{tabular}
  \vspace{-5pt}
\end{table}

%\begin{comment}
%end{comment}

\newpage
\clearpage
\newpage

\section*{Appendix C.  OSM / SpaceNet Model Comparison}

\begin{figure}[h]
\vspace{-5pt}
\begin{center}
\setlength{\tabcolsep}{0.3em}
\begin{tabular}{c}
\vspace{-5pt}

\subfloat {\includegraphics[width=0.98\linewidth]{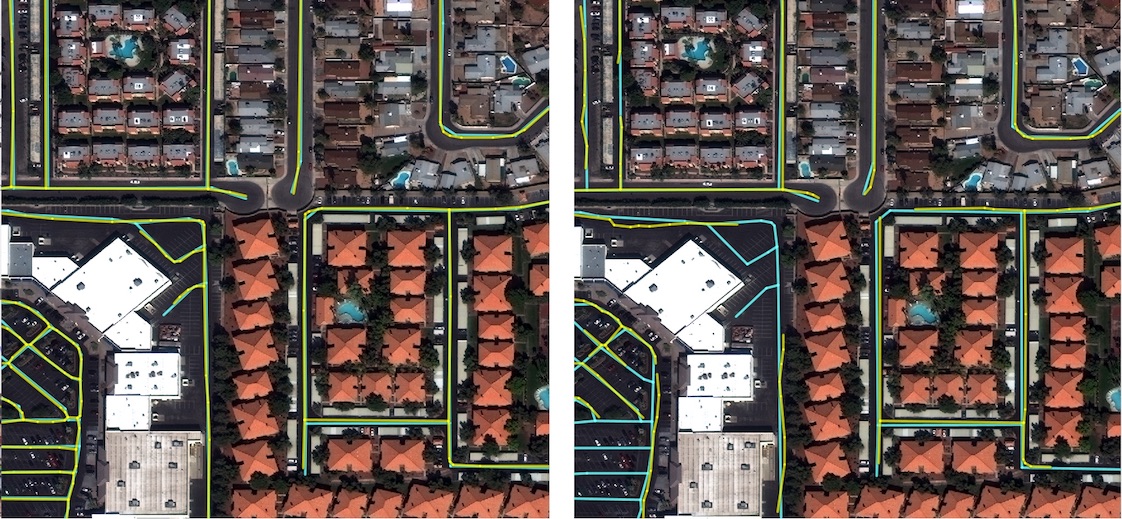}} \\
\subfloat {\includegraphics[width=0.98\linewidth]{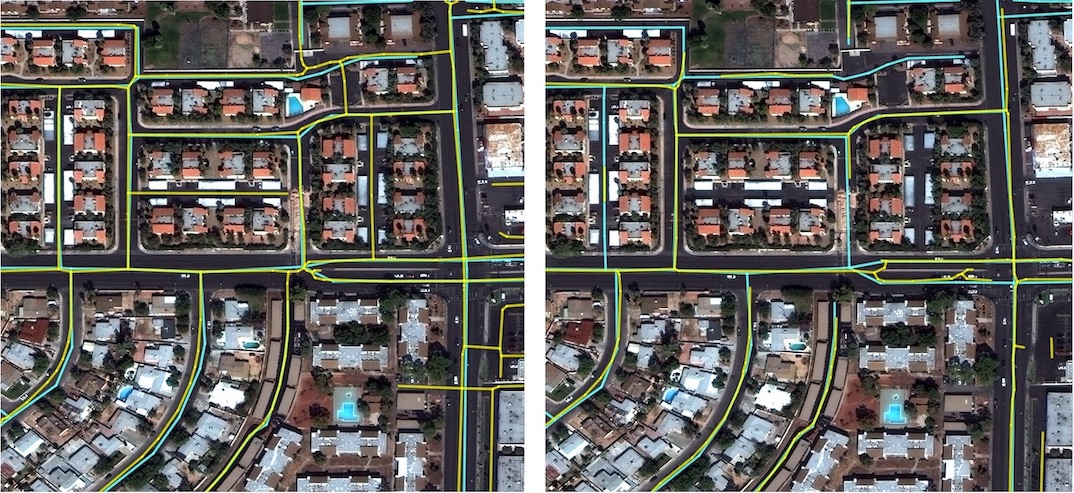}} \\

\end{tabular}

 \caption{\textbf{SpaceNet compared to OSM.} Road predictions (yellow) and ground truth SpaceNet labels (blue) for a sample image chips, with SpaceNet models on the left and OSM-trained models on the right.  
 Top: SpaceNet model predictions (left) score APLS$_{\rm{length}} = 0.60$,  while OSM model predictions (right)  yield APLS$_{\rm{length}} = 0.48$.
 Bottom: SpaceNet model predictions (left) score APLS$_{\rm{length}} = 0.92$,  while OSM model predictions (right)  yield APLS$_{\rm{length}} = 0.37$.
 }
 \label{fig:blah} 
\end{center}
\vspace{-10pt}
\end{figure}

Figure \ref{fig:blah} displays comparisons of models trained on OSM data and SpaceNet data.

\newpage

\section*{Appendix D. CRESIv2 Road Speed Plots}

\begin{figure}[h]
\vspace{-5pt}
\begin{center}
\setlength{\tabcolsep}{0.2em}
\begin{tabular}{c}
%\vspace{-5pt}

%\subfloat {\includegraphics[width=0.98\linewidth]{khartoum2.jpg}} \\
%\vspace{2pt}
\subfloat {\includegraphics[width=0.98\linewidth]{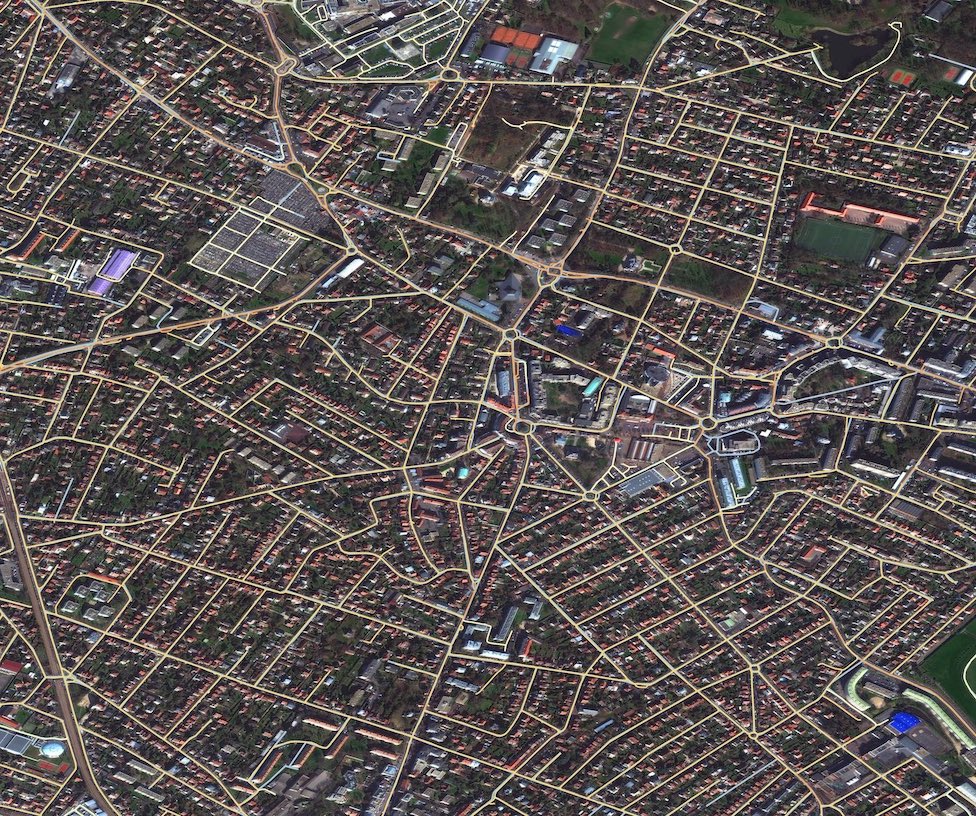}} \\
\vspace{2pt}
\subfloat {\includegraphics[width=0.98\linewidth]{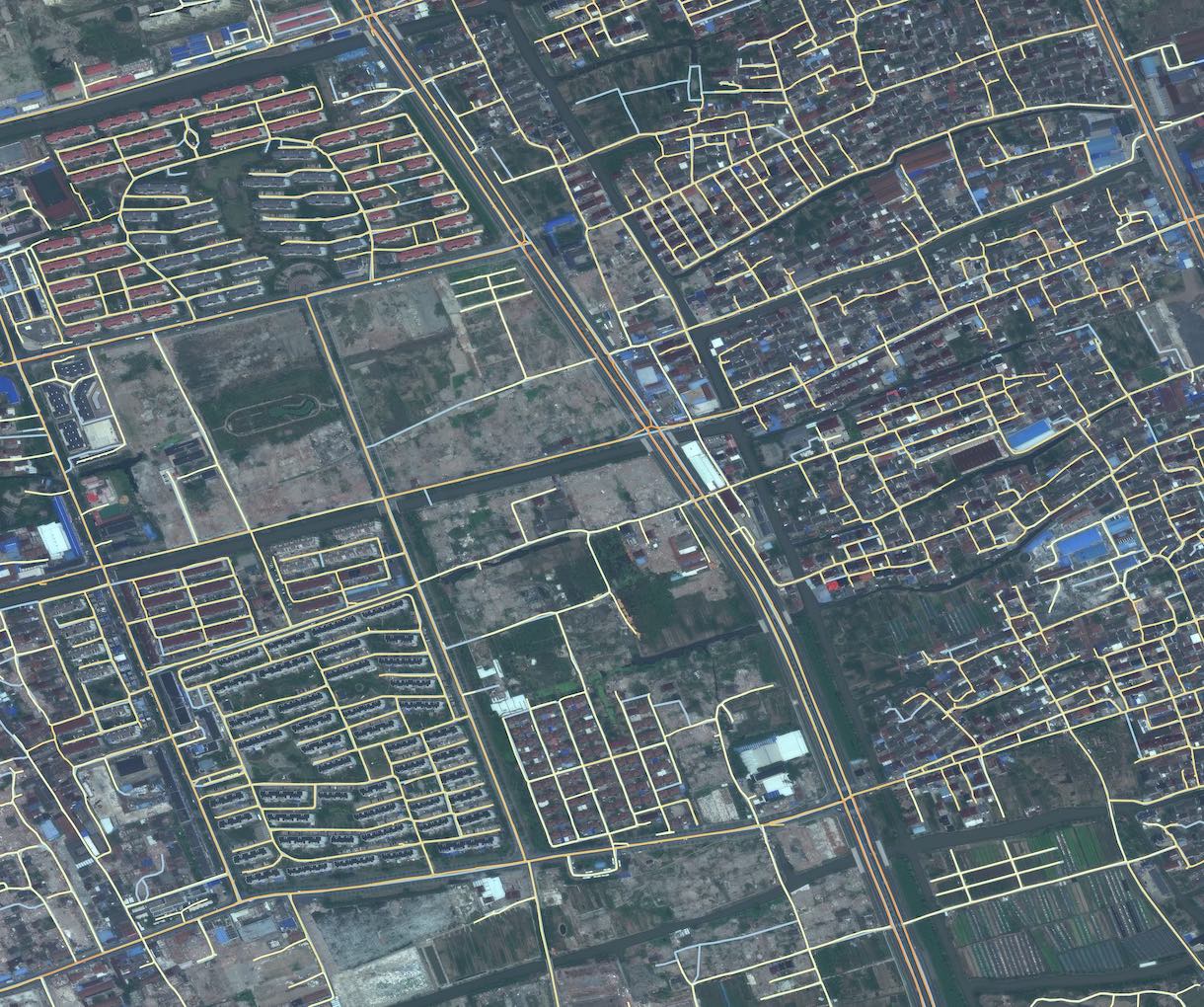}} \\
\vspace{-10pt}

\end{tabular}
\caption{\textbf{Road speed.} Output of CRESIv2 inference as applied to large SpaceNet test regions (from top: Paris, Shanghai). 
Roads are colored by inferred speed limit, from yellow
 (20 mph) 
%($\leq 20$ mph) 
to red (65 mph), with ground truth in gray. }

 \label{fig:comp2}
\end{center}
\vspace{-6pt}
\end{figure}

\begin{comment}
\begin{figure}[h]
\vspace{-5pt}
  \centering
  \includegraphics[width=0.98\linewidth]{khartoum2.jpg}
\caption{\textbf{CRESIv2 road speed.} Output of CRESIv2 inference as applied to a portion of the SpaceNet Khartoum test region. Roads are colored by inferred speed limit, from yellow (20 mph) to red (65 mph).}
\label{fig:speed1}
\vspace{-5pt}
\end{figure}

\begin{figure}[h]
\vspace{-5pt}
  \centering
  \includegraphics[width=0.98\linewidth]{paris0.jpg}
\caption{\textbf{CRESIv2 road speed.} Output of CRESIv2 inference as applied to a portion of the SpaceNet Paris test region. Roads are colored by inferred speed limit, from yellow (20 mph) to red (65 mph).}
\label{fig:speed2}
\vspace{-5pt}
\end{figure}

\begin{figure}%[h]
\vspace{-5pt}
  \centering
  \includegraphics[width=0.98\linewidth]{shanghai1.jpg}
\caption{\textbf{CRESIv2 road speed.} Output of CRESIv2 inference as applied to a portion of the SpaceNet Shanghai est region. Roads are colored by inferred speed limit, from yellow (20 mph) to red (65 mph).}
\label{fig:speed3}
\vspace{-5pt}
\end{figure}

\end{comment}

%\clearpage
\newpage

\section*{Appendix E. CRESIv2 / RoadTracer Visual Comparison}

\begin{figure}[h]
\vspace{-5pt}
\begin{center}
\setlength{\tabcolsep}{0.2em}
\begin{tabular}{cc}
\vspace{-5pt}

\subfloat {\includegraphics[width=0.48\linewidth]{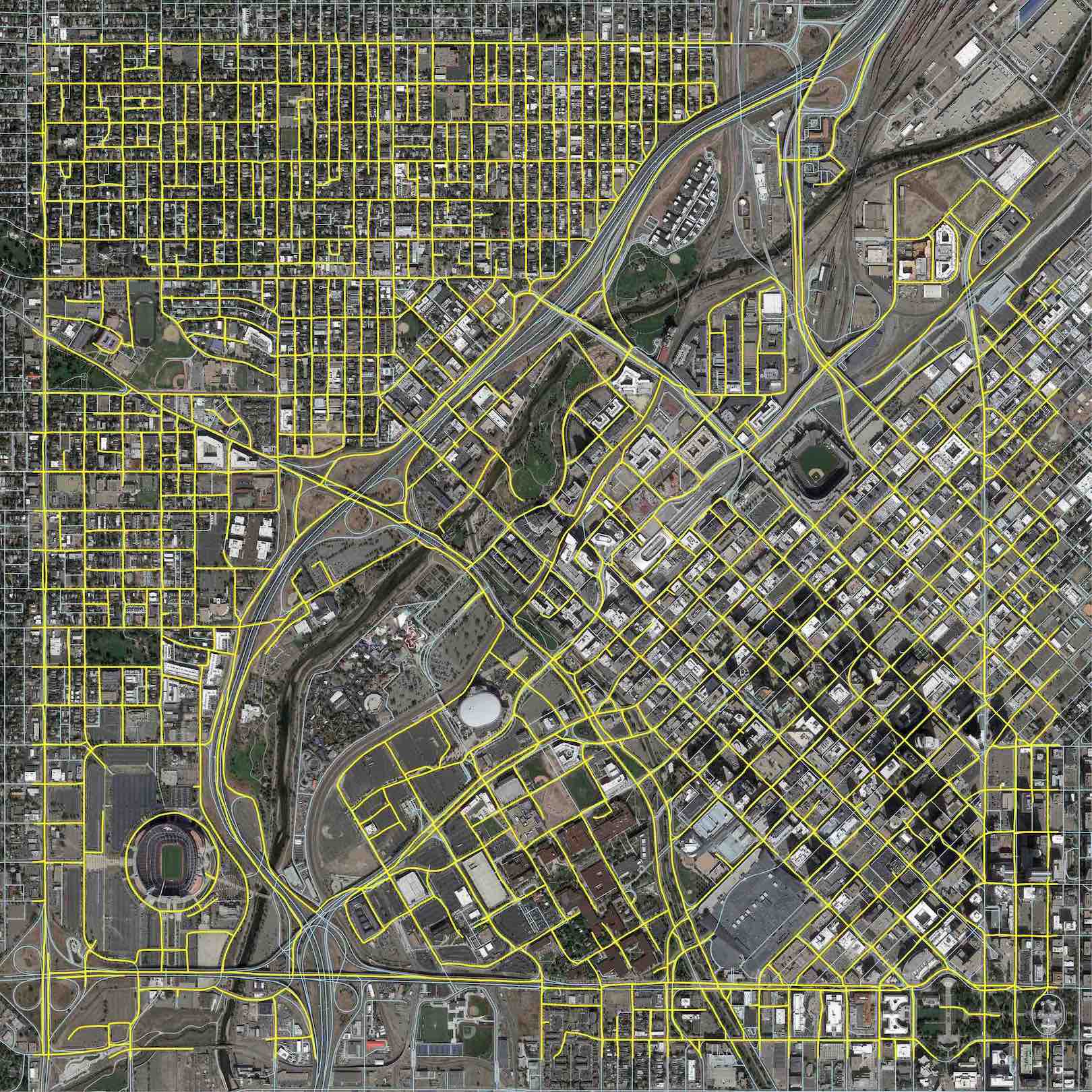}} &
\subfloat {\includegraphics[width=0.48\linewidth]{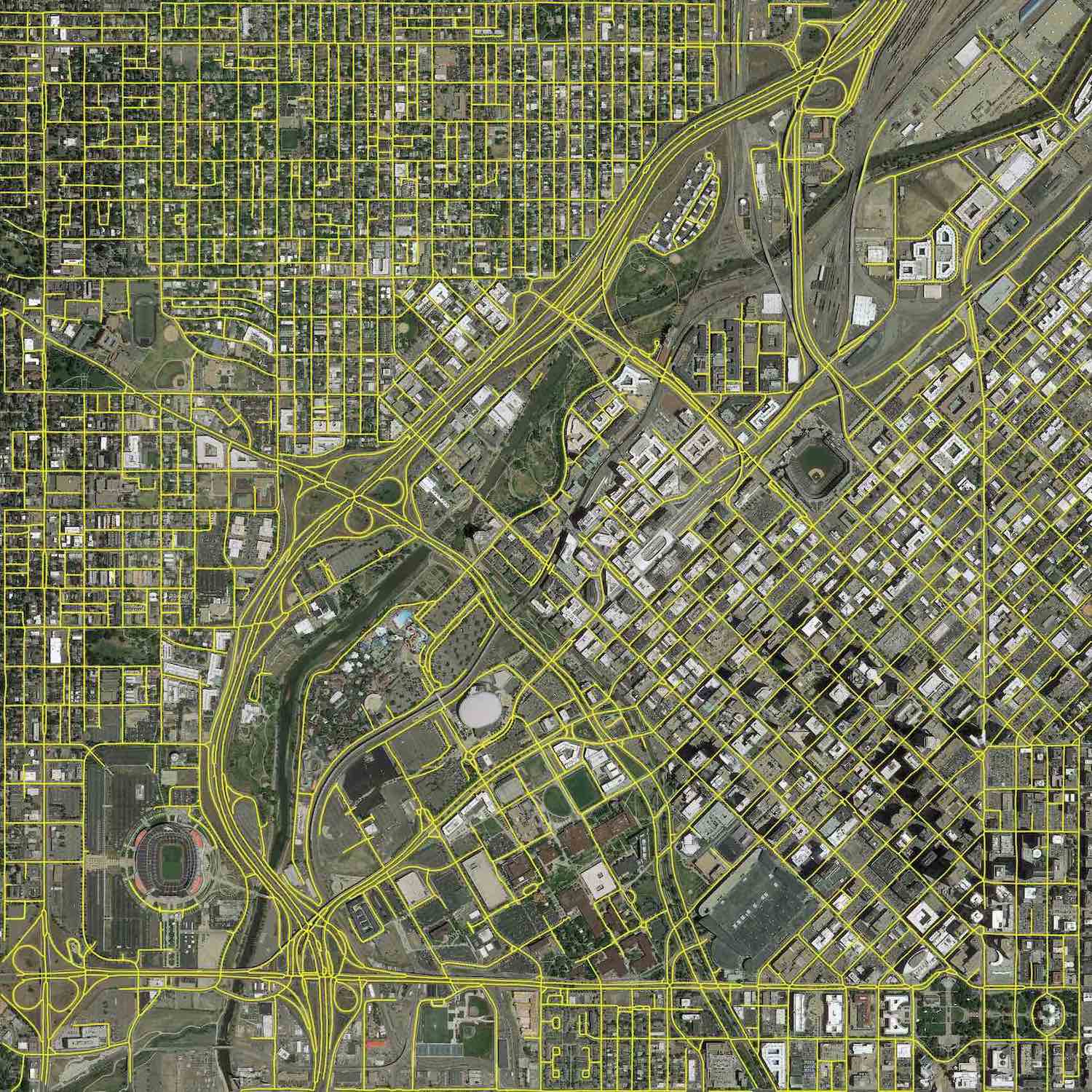}} \\
\vspace{-5pt}

%\subfloat {\includegraphics[width=0.48\linewidth]{pittsburgh_roadtracer.jpg}} &
%\subfloat {\includegraphics[width=0.48\linewidth]{pittsburgh_ox_plot.jpg}} \\
%\vspace{-5pt}

%\subfloat {\includegraphics[width=0.48\linewidth]{toronto_roadtracer.jpg}} &
%\subfloat {\includegraphics[width=0.48\linewidth]{toronto_ox_plot.jpg}} \\
%\vspace{-5pt}

\subfloat {\includegraphics[width=0.48\linewidth]{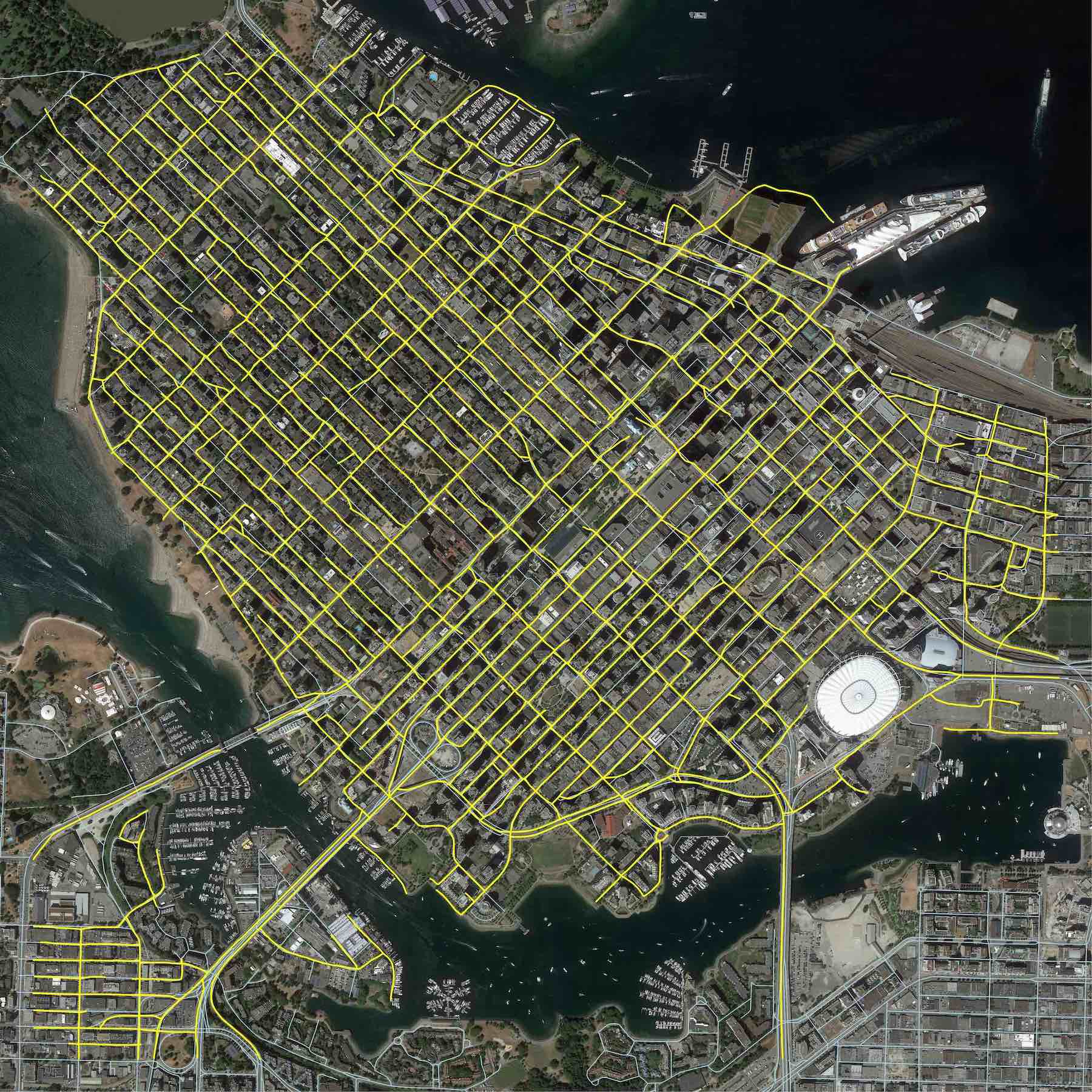}} &
\subfloat {\includegraphics[width=0.48\linewidth]{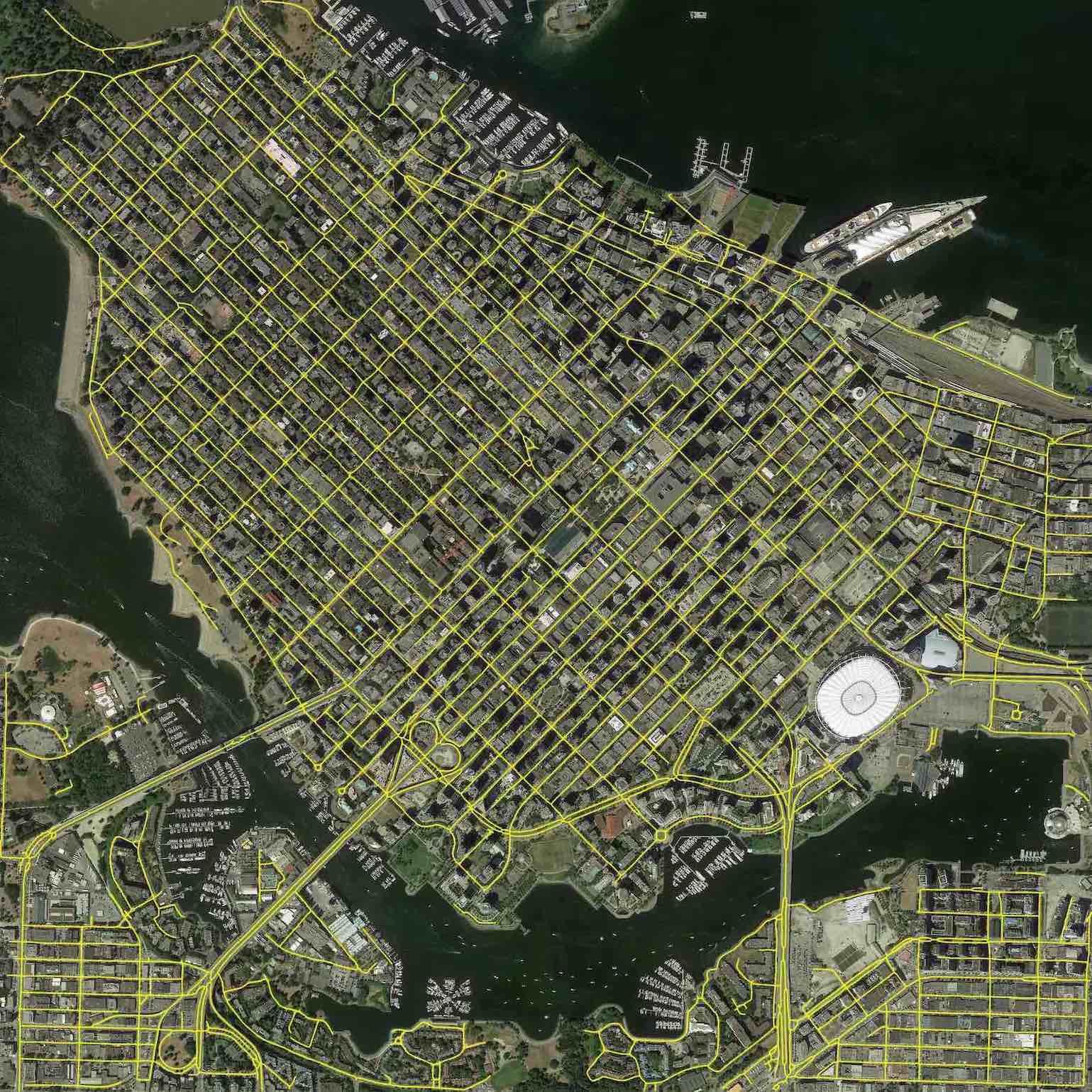}} \\
\vspace{-5pt}

\end{tabular}

 %\caption{\textbf{New York City Performance.} (a) RoadTracer prediction \cite{roadtracer_ims} (b) Our CRESIv2 prediction over the same area. }
 \caption{\textbf{RoadTracer / CRESIv2.} Performance comparison between RoadTracer (left column, OSM labels in gray, predictions in yellow  \cite{roadtracer_ims}) and CRESIv2 (right column, predictions in yellow) for various cities.  
 From top:
 Denver, 
%  Pittsburgh, 
% Toronto, 
 Vancouver.}

 \label{fig:comp2}
\end{center}
\vspace{-6pt}
\end{figure}

%\end{comment}
%%%%%%%%%%

\newpage

\section*{Appendix F.  RoadTracer / CRESIv2 / DeepRoadMapper Zooms}

\begin{figure}[h!]
\begin{center}
\includegraphics[width=0.99\linewidth]{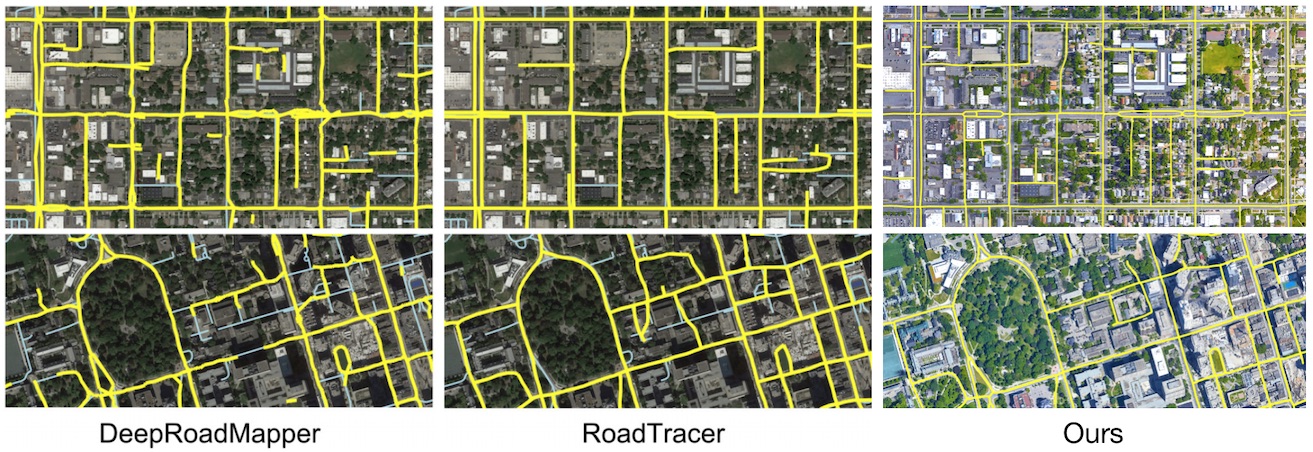}
\end{center}
\caption{Qualitative comparison of three methods over various cities, (see Figure 10 of \cite{roadtracer}).}
\label{fig:rt}
\end{figure}

A qualitative comparison of three methods over various cities is shown in Figure \ref{fig:rt}.

\end{document}